\documentclass[sw]{iosart2x}
\usepackage{dsfont}
\usepackage{subcaption}
\usepackage{multirow}
\usepackage[dvipsnames]{xcolor}
\let\oldhat\hat
\renewcommand{\vec}[1]{\mathbf{#1}}
\renewcommand{\hat}[1]{\oldhat{\mathbf{#1}}}

\pubyear{2021}
\volume{Deep Learning and Knowledge Graphs}
\firstpage{1}
\lastpage{1}

\begin{document}
\makeatletter
\let\put@numberlines@box\relax
\makeatother
\begin{frontmatter}

\title{A Survey on Visual Transfer Learning using Knowledge Graphs}
\runtitle{A Survey on Visual Transfer Learning using Knowledge Graphs}


\author[A,B]{\inits{S.}\fnms{Sebastian} \snm{Monka}\ead[label=e1]{sebastian.monka@de.bosch.com}}
and
\author[A]{\inits{L.}\fnms{Lavdim} \snm{Halilaj}\ead[label=e2]{lavdim.halilaj@de.bosch.com}}
and
\author[B]{\inits{A.}\fnms{Achim} \snm{Rettinger}\ead[label=e3]{rettinger@uni-trier.de}}
\address[A]{Corporate Research, \orgname{Robert Bosch GmbH},
Renningen, \cny{Germany}\printead[presep={\\}]{e1,e2}}
\address[B]{Computer Sciences, \orgname{Trier University},
Trier, \cny{Germany}\printead[presep={\\}]{e3}}


\begin{abstract}
The information perceived via visual observations of real-world phenomena is unstructured and complex.
\emph{Computer vision} (CV) is the field of research that attempts to make use of that information.
Recent approaches of CV utilize \emph{deep learning} (DL) methods as they perform quite well if training and testing domains follow the same underlying data distribution.
However, it has been shown that minor variations in the images that occur when these methods are used in the real world can lead to unpredictable and catastrophic errors.
Transfer learning is the area of machine learning that tries to prevent these errors.
Especially, approaches that augment image data using auxiliary knowledge encoded in language embeddings or \emph{knowledge graphs} (KGs) have achieved promising results in recent years.
This survey focuses on visual transfer learning approaches using KGs, as we believe that KGs are well suited to store and represent any kind of auxiliary knowledge.
KGs can represent auxiliary knowledge either in an underlying graph-structured schema or in a vector-based \emph{knowledge graph embedding}.
Intending to enable the reader to solve visual transfer learning problems with the help of specific KG-DL configurations we start with a description of relevant modeling structures of a KG of various expressions, such as directed labeled graphs, hypergraphs, and hyper-relational graphs.
We explain the notion of feature extractor, while specifically referring to visual and semantic features.
We provide a broad overview of \emph{knowledge graph embedding methods} and describe several joint training objectives suitable to combine them with high dimensional visual embeddings.
The main section introduces four different categories on how a KG can be combined with a DL pipeline:
1) Knowledge Graph as a Reviewer; 
2) Knowledge Graph as a Trainee; 
3) Knowledge Graph as a Trainer;
and 4) Knowledge Graph as a Peer.
To help researchers find meaningful evaluation benchmarks, we provide an overview of generic KGs and a set of image processing datasets and benchmarks that include various types of auxiliary knowledge.
Last, we summarize related surveys and give an outlook about challenges and open issues for future research.
\end{abstract}


\end{frontmatter}








\section{Introduction}


\emph{Deep learning} (DL) as a \emph{machine learning} (ML) technique is broadly used to successfully solve \emph{computer vision} (CV) tasks.
Their main strength is their ability to find complex underlying features in a given set of images.
A common method for training a \emph{deep neural network} (DNN) is to minimize the \emph{cross-entropy} (CE) loss, which is equivalent to maximizing the negative log-likelihood between the empirical distribution of the training set and the probability distribution defined by the model. 
This relies on the \emph{independent and identically distributed} (i.i.d.) assumptions as underlying rules of basic ML, which state that the examples in each dataset are independent of each other, that train and test set are identically distributed and drawn from the same probability distribution~\cite{DBLP:books/daglib/0040158}.
However, if the train and test domains follow different image distributions the i.i.d. assumptions are violated, and DL leads to unpredictable and poor results~\cite{DBLP:conf/icann/TanSKZYL18}.
This has been demonstrated by using adversarially constructed examples~\cite{DBLP:journals/corr/GoodfellowSS14} or variations in the test images such as noise, blur, and JPEG compression~\cite{DBLP:conf/iclr/HendrycksD19}.
Moreover, authors in~\cite{DBLP:journals/corr/abs-2011-03395} even claim that any standard DNN suffers from such an unpredictable distribution shift when it is deployed in the real world.

Transfer learning is the area of machine learning that groups approaches dealing with such an unpredictable distribution shift~\cite{DBLP:journals/corr/abs-2011-03395}.
Most of the transfer learning approaches try to solve the problem by inducing a bias into the DNN to overcome data issues.
Especially, approaches that extend image data using auxiliary knowledge encoded in language embeddings or \emph{knowledge graphs} (KGs) have achieved promising results in recent years.
Due to Larochelle et al.~\cite{DBLP:conf/aaai/LarochelleEB08} auxiliary knowledge is not only important to solve transfer learning problems, but also an opportunity to influence the way an ML model learns from unstructured data.

In this survey, we focus on visual transfer learning approaches using KGs, as we believe that KGs are well suited to store and represent any kind of auxiliary knowledge.
The auxiliary knowledge encoded in an underlying graph-structured schema can then be converted to a vector-based \emph{knowledge graph embedding} ($h_s$).
The ability to transform the graph-based knowledge into the vector space enables the application of linear operations thus its use in combination with DNNs.
A commonly used method for introducing auxiliary knowledge is to use a joint training objective that combines the semantic embedding $h_s$ with the visual embedding $h_v$.
In the scope of the survey we introduce three distinct types of joint embeddings:
a) A semantic-visual embedding $h_{s,v}$, where semantic data is embedded using $h_v$ as an objective;
b) A visual-semantic embedding $h_{v,s}$, where visual data is embedded using $h_s$ as an objective; and
c) A hybrid embedding $h_{h}$, where both semantic and visual data are embedded using a combination of $h_s$ and $h_v$ as an objective.

Our main contributions in this survey are listed in the following:
\begin{itemize}
\renewcommand{\labelitemi}{$\bullet$}
    \item A categorization of visual transfer learning approaches using KGs according to four distinct ways a KG can be combined with a DL pipeline.
    \item A description of generic KGs and relevant datasets and benchmarks for visual transfer learning using KGs for CV tasks.
    \item A comprehensive summary of the existing surveys on visual transfer learning using auxiliary knowledge.
    \item An analysis of research gaps in the area of visual transfer learning using KGs which can be used as a basis for future research.
\end{itemize}

The remainder of this paper is structured as follows: Section~\ref{sec:Methodology} provides an overview of the methodology followed to conduct the survey.
In Section~\ref{sec:Background} we introduce the term visual transfer learning.
In addition, we outline different types of modeling structures of knowledge graphs such as directed labeled graphs, hypergraphs, and hyper-relational graphs.
We explain the notion of features extractor, specifically referring to visual and semantic features.
Further, we describe the term knowledge graph embedding and provide a brief categorization of KGE-Methods concerning different supervision and input types.
Several joint training objectives suitable to combine semantic embeddings with visual embeddings are described.
The main section, Section~\ref{sec:Visual Transfer Learning using Knowledge Graphs} introduces four different categories on how a KG can be combined with a DL pipeline:\\
1) \emph{Knowledge Graph as a Reviewer} - where the KG is used for post-validation of a visual model;\\
2) \emph{Knowledge Graph as a Trainee}, where the KG is embedded into $h_{s,v}$ using $h_v$ as objective;\\
3) \emph{Knowledge Graph as a Trainer}, where the KG with $h_s$ is used as an objective to embedd images into $h_{v,s}$; and\\
4) \emph{Knowledge Graph as a Peer}, where the KG with $h_s$ is combined with $h_v$ to suit as objective that embedds both the KG and images into $h_h$.\\
Since KGE-Methods have only recently entered the field of visual transfer learning, we also list related methods forming $h_s$ based on other types of auxiliary knowledge in categories 2), 3), and 4).
Other types of auxiliary knowledge are language descriptions or class attributes so that their semantic features extractor $f_{s}(\cdot)$ differs in the type of input, but not in its architecture.
Furthermore, in Section~\ref{sec:Datasets and Benchmarks} we provide an overview of generic KGs, several datasets and benchmarks using various types of auxiliary knowledge, like attributes, textual descriptions, or graphs.
In Section~\ref{sec:Related Surveys} we summarize related surveys in the field of visual transfer learning and knowledge-based ML.
Section~\ref{sec:Challenges and Open Issues} gives an outlook about challenges and open issues in the field of visual transfer learning using knowledge graphs.
Finally, Section~\ref{sec:conclusion} provides a discussion and a conclusion as well as an outlook of future directions on the field.




\section{Methodology}
\label{sec:Methodology}

Our objective is to provide a comprehensive overview of how KGs can be used in combination with DL to solve visual transfer learning tasks.
To ensure the quality of the outcome, we followed the process proposed by Petersen et. al~\cite{DBLP:conf/ease/PetersenFMM08,DBLP:journals/infsof/PetersenVK15} and conducted an initial search on five scholarly indexing services.
We applied inclusion and exclusion criteria on our findings and extended them based on the snowballing approach~\cite{DBLP:conf/ease/Wohlin14}.

\subsection{Research Questions}

The combination of visual and semantic data seems to be a promising direction to build models that can cope with the diversity of the real world.
However, some major challenges and questions arise when combining these modalities.
\begin{itemize}
    \item \textbf{RQ1} - How can a knowledge graph be combined with a deep learning pipeline?
    \item \textbf{RQ2} - What are the properties of the respective combinations?
    \item \textbf{RQ3} - Which knowledge graphs already exist, that can be used as auxiliary knowledge?
    \item \textbf{RQ4} - What datasets exist, that can be used in the combination with auxiliary knowledge to evaluate visual transfer learning?
\end{itemize}
\textbf{RQ1} and \textbf{RQ2} are answered in Section~\ref{sec:Visual Transfer Learning using Knowledge Graphs}, where we categorize and discuss visual transfer learning approaches based on how the KG is combined with the DL pipeline.
\textbf{RQ3} and \textbf{RQ4} are answered in Section~\ref{sec:Datasets and Benchmarks}, where we summarize available KGs, datasets, and benchmarks that will help to compare approaches of the field of visual transfer learning using KGs.

\subsection{Literature Search}
To collect relevant literature, we define a search string using the following strategy:

\begin{itemize}
    \item Extract major terms from research questions.
    \item Use synonyms and alternative terms.
    \item Combine using \emph{OR} to include synonyms and alternative terms.
    \item Combine using \emph{AND} to join the key terms.
\end{itemize}

As a result, the following major terms related to the concepts are derived: 
Knowledge Graph, 
Visual Transfer Learning, 
and connect them by a Boolean AND operation. 
Each term contains a set of keywords related to the respective concept, connected by a Boolean OR operation.
Therefore, the initial search string was as follows:
\textbf{(("Knowledge Graph" OR "Knowledge Graph Embedding" OR "Semantic Embedding") AND ("Visual Transfer Learning" OR "Transfer Learning" OR "Zero-shot Learning" OR "Deep Learning" OR "Computer Vision"))}

For the primary search process we used five scholarly indexing services: 
Google Scholar\footnote{https://scholar.google.com}, 
IEEE Xplore\footnote{https://ieeexplore.ieee.org}, 
ACM Digital Library\footnote{https://dl.acm.org}, Scopus\footnote{https://www.scopus.com}, 
and DBLP\footnote{https://dblp.uni-trier.de}.

\subsection{Literature Selection and Quality Assessment}
After the literature search we included literature based on the following criteria:
\begin{itemize}
    \item Studies using visual features.
    \item Studies using auxiliary knowledge.
\end{itemize}

Further, we excluded literature based on the following criteria:
\begin{itemize}
    \item News articles.
    \item Non-English studies.
    \item Non-public available studies.
    \item Duplicate studies.
\end{itemize}

We reduced the amount of 16,200 studies after applying the inclusion and exclusion criteria on title and abstract to 17 relevant surveys and 164 studies ($1.12\%$)
During full-text reading, it became obvious that further articles should be removed as they were not in the scope based on the inclusion and exclusion criteria. 
The remaining articles (106) were used to conduct backward snowball sampling~\cite{DBLP:conf/ease/Wohlin14}, which led to 22 additional studies.
On the set of 128 primary studies we conducted a quality assessment on the following questions:
\begin{itemize}
    \item Does the study provide a solid assessment?
    \item Are the results plausible?
\end{itemize}
Thus, we were able to reduce the number of studies to 124. 
These studies provide the basis for the survey and serve to answer the formulated research questions.

\section{Background}
\label{sec:Background}

This section briefly introduces the term visual transfer learning, describes the fundamentals of KGs, feature extractors, knowledge graph embeddings, and joint training objectives in the context of this survey.

\subsection{Visual Transfer Learning}
\label{ssec:Visual Transfer Learning}

Visual transfer learning is presented in \cite{DBLP:conf/emnlp/RuderP17} as follows: 
\emph{Given a source domain $D_S$ with input data $X_S$, a corresponding source task $T_S$ with labels $Y_S$, as well as a target domain $D_T$ with input data $X_T$ and a target task $T_T$ with labels $Y_T$, the objective of visual transfer learning is to learn the target conditional probability distribution $P_T (Y_T | X_T )$ with the information gained from $D_S$ and $T_S$ where $D_S \neq D_T$ or $T_S \neq T_T$}.
\paragraph{Zero-Shot Learning}
is a visual transfer learning task with labeled source domain data and unlabeled target domain data.
Zero-shot learning aims to extract implicit knowledge of the classes in the source domain task $T_S$ and transfers this knowledge to unknown classes of the target domain task $T_T$~\cite{DBLP:journals/tkde/PanY10}.
If zero-shot learning has access to an additional set of labeled target data $X_T$, the task is called few-shot learning.

\paragraph{Domain Generalization}
is a visual transfer learning task with access to labeled source domain data and unlabeled target domain data.
Domain generalization aims to extract implicit knowledge of the source domain $D_S$ and transfer this knowledge to an unknown target domain $D_T$~\cite{DBLP:conf/nips/BlanchardLS11,DBLP:conf/icml/MuandetBS13}.
If domain generalization has access to an additional set of labeled target data $X_T$, the task is called domain adaptation.

\subsection{Knowledge Graph}
\label{ssec:Knowledge Graph}

Knowledge is the awareness, understanding, or information for a phenomenon or a subject that has been obtained by observations or study\footnote{https://dictionary.cambridge.org/dictionary/english/knowledge}.
It can be either implicit or explicit and stored and represented in different ways.
Explicit knowledge is the type of knowledge that can be easily interpreted, organized, managed, and transmitted to others. 
Implicit knowledge is the form of knowledge that is gathered through observations and activities of everyday life.
Using various modeling techniques, complex explicit knowledge can be formally represented in KGs.
On the other hand, a common method for gathering implicit knowledge is to use feature extraction methods, that learn latent knowledge representations, e.g. visual or semantic embeddings, from observations~\cite{DBLP:books/daglib/0040158}.

There exist many ways for expressing, representing, and storing knowledge.
In this survey, we focus on KGs, a structured representation of facts, consisting of entities, relationships, and semantic descriptions. 
A comprehensive definition is given by the authors of~\cite{DBLP:journals/corr/abs-2003-02320} where a KG is defined as \emph{a graph of data with the objective of accumulating and conveying real-world knowledge, where entities are represented by nodes and relationships between entities are represented by edges}.
Knowledge can be expressed in a factual triple in the form of (head, relation, tail).
In its most basic form, we see 
a KG as a set of triples $G = {H, R, T}$, where $H$ is a set of entities, $T \subseteq E \times L $, is a set of entities and literal values and $R$, set of relationships which connects $H$ and $R$.

A graph model is a model which structures the data, including its schema and/or instances in form of graphs, and the data manipulation is realized by graph-based operations and adequate integrity
constraints~\cite{DBLP:books/sp/18/AnglesG18}.
Each graph model has its formal definition based on the mathematical foundation, which can vary according to different characteristics, for instance, directed vs undirected, labeled vs unlabeled, etc.
The most basic model is composed of labeled nodes and edges, easy to comprehend but inappropriate to encapsulate multidimensional information.
Other graph models allow for the representation of information utilizing complex relationships in the form of hypernodes or hyperedges. 
In the following, we discuss three common graph models that are used in practice to represent data graphs.

\begin{table*}[htb]
 \caption{\textbf{Various Graph Models}. Three common graph models used as underlying structure for knowledge representation in KGs: 1) Directed Labeled Graphs; 2) Hypergraphs; and 3) Hyper-relational Graphs.}
  \label{tab:graphModels}
	\centering
	\begin{tabular}{|p{1.5cm}|p{4.3cm}|p{4.3cm}|p{4.3cm}|}
	\hline
      & \textbf{Directed Labeled Graphs} & \textbf{Hypergraphs} & 
      \textbf{Hyper-Relational Graphs}
    \\ \hline
    Nodes and Literals & 
    - Real-world and abstract entities \newline
    - Entity's attribute value 
     & 
       - Real-world and abstract entities \newline
       - Entity's attribute value 
     & 
       - Real-world and abstract entities \newline
       - Entity's attribute value 
    \\
   \hline
   Relationships & 
      - Binary relations between entities \newline
      - Relations between an entity and its attribute's values
    &
      - Binary relations between entities \newline
      - Relations between an entity and its attribute's values \newline
      - Many-to-many relations between entities (Hyperedge)
 &
     - Binary relations between entities \newline
     - Relations between an entity and its attribute's values \newline
     - Additional information encoded in relationship (Hyper-relation)
     
\\
  \hline 
    Semantics & 
    Connect two nodes & 
    Connect an arbitrary set of nodes & 
    Connect two nodes with additional contextual information\\
  \hline
  Example &
  \includegraphics[width=4.3cm]{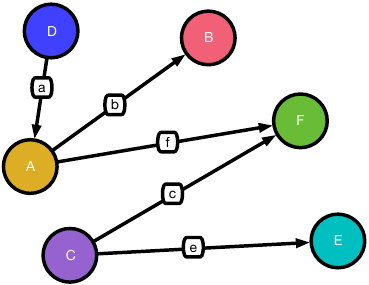} &
  \includegraphics[width=4.3cm]{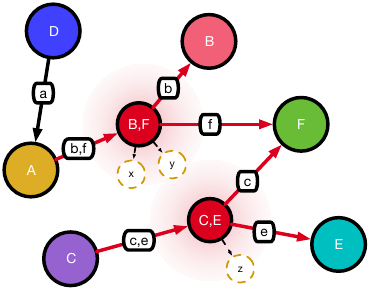} &
  \includegraphics[width=4.3cm]{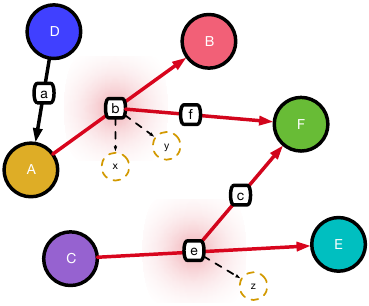}\\
  \hline

	\end{tabular}
\end{table*}

\paragraph{Directed Labeled Graphs:}

A directed labeled graph is comprised of a set of nodes and a set of edges connecting those nodes, labeled based on a specific vocabulary~\cite{DBLP:books/sp/18/AnglesG18}.

The direction of the edge of two paired nodes is important, which clearly distinguishes between the start node and the end node.
This intuitively enables the organization of information via the utilization of binary relationships. 

\paragraph{Hypergraphs:}

Hypergraphs extend the definition of binary edges by allowing the modeling of multiple and complex relationships~\cite{DBLP:books/sp/18/AnglesG18}.

On the other hand, hypernodes modularize the notion of node, by allowing nesting graphs inside nodes.
In addition, the notion of a hyperedge enables the definition of n-ary relations between different concepts.

\paragraph{Hyper-Relational Graphs:}

A hyper-relational graph is also a labeled directed multigraph where each node and edge might have several associated key-value pairs \cite{DBLP:journals/access/AnglesTT20}.

Internally, nodes and edges are annotated according to a chosen vocabulary and have unique identifiers, making them a flexible and powerful form of modeling for graph analysis with weighted edges. 

Table~\ref{tab:graphModels} illustrates the three graph models mentioned above with some corresponding examples.
A KG can be based on any such graph model utilizing nodes and edges as a fundamental modeling form.

\subsection{Feature Extractor}

A feature extractor is a transformation function from higher dimensional into lower dimensional vector space, including a vast variety of dimensionality reduction methods~\cite{DBLP:books/lib/Bishop07,DBLP:journals/pr/WangP03}.

Since it has been shown that most downstream tasks can be solved better on a reduced dimensionality, feature extractors are also a fundamental building block of modern systems working on visual and semantic data.

However, more and more conventional feature extraction methods have been replaced with DNNs.
A DNN is an artificial \emph{neural network} (NN) with multiple layers between the input and output layers, having the ability to automatically extract lower dimensional features from the input data~\cite{hinton2006reducing,DBLP:conf/cvpr/KavukcuogluRFL09}.

\begin{figure}[h]
\includegraphics[width=0.45\textwidth]{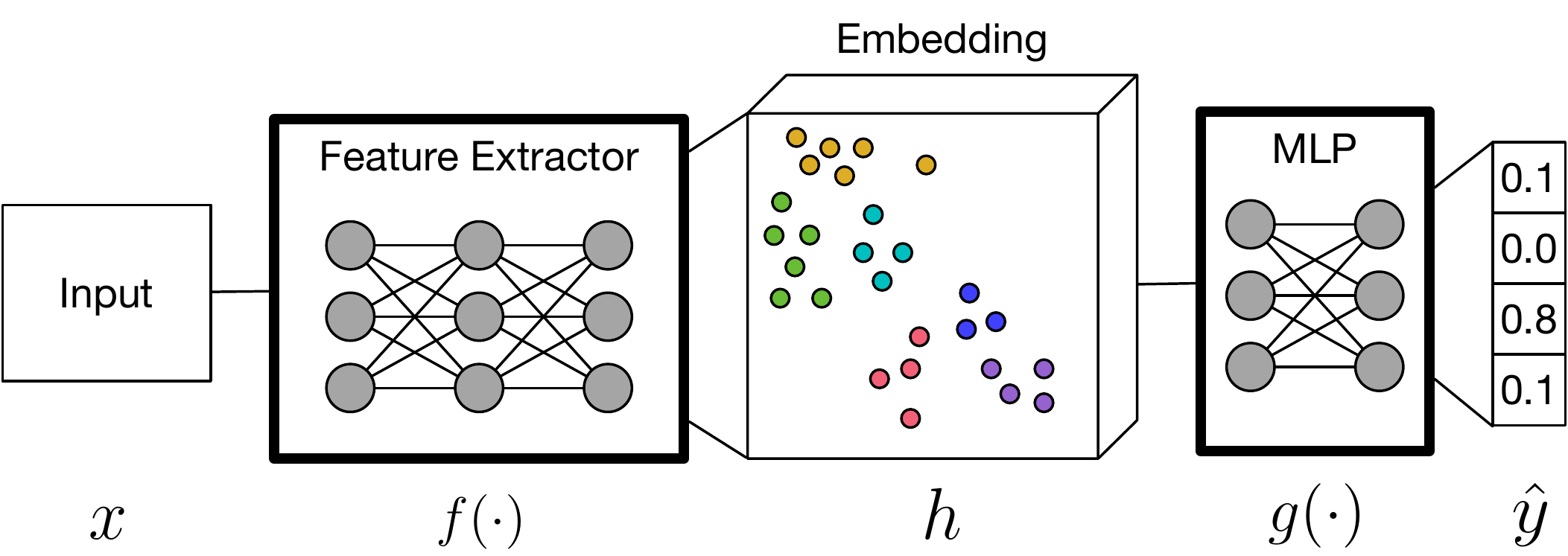}
\caption{A DNN that takes $\vec{x}$ as input and predicts $\hat{\vec{y}}$ can be decoupled into a feature extractor $f(\cdot)$ with its embedding space $\vec{h}$ and a prediction task $g(\cdot)$.}
\label{fig:DNN}
\end{figure}

As depicted in Figure~\ref{fig:DNN}, a DNN can be decoupled in a feature extractor $f(\cdot)$, with its embedding space $\vec{h}$ and a prediction task $g(\cdot)$, expressing the function
\begin{equation}
\hat{\vec{y}} = g(f(\vec{x})) \text{, with } f(x) = \vec{h}.
\label{eq:visual features extractor}
\end{equation}

There are different architectures of DNNs, but they always consist of the same components: neurons, synapses, weights, biases, and functions~\cite{DBLP:books/daglib/0040158}.
The most common architectures that build a DNN are \emph{multilayer perceptrons} (MLP), \emph{convolutional neural networks} (CNN), \emph{recurrent neural networks} (RNN), and \emph{transformer models}.
Each architecture has its advantages and is therefore preferred for a particular type of input data and particular task~\cite{DBLP:books/daglib/0040158}.

Whereas, DNNs are usually trained end-to-end resulting in a task-dependent embedding space $\vec{h}$, more recently, attempts have been made to independently pre-train the feature extractor that it can be applied to several visual transfer learning and downstream tasks~\cite{DBLP:conf/nips/ChenKSNH20}.

\begin{figure*}[t]
        \begin{subfigure}{0.49\textwidth}
    \centering
    \includegraphics[width=0.9\textwidth]{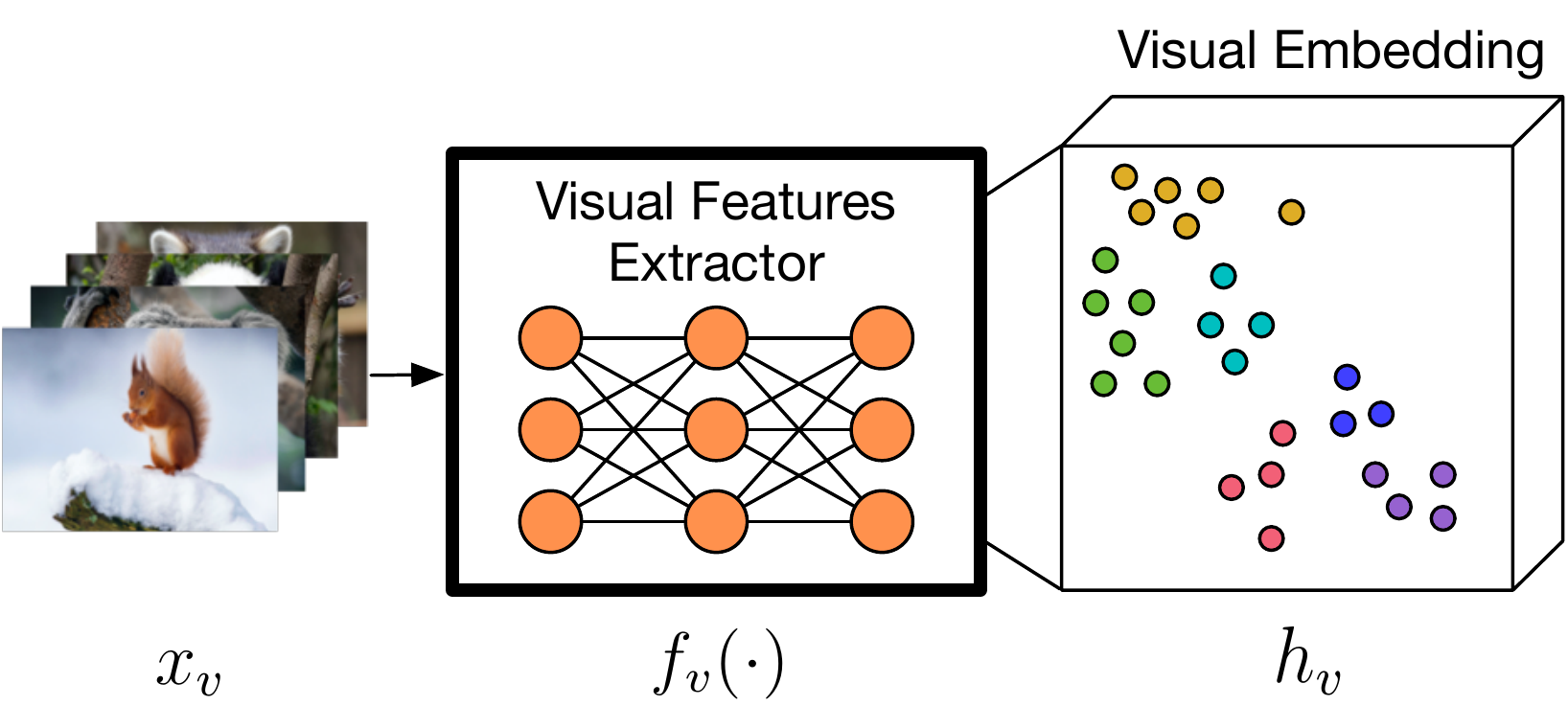}
    \caption{Visual features extractor}
    \label{fig:visual features extractor}
    \end{subfigure}
    ~
    \begin{subfigure}{0.49\textwidth}
    \centering
    \includegraphics[width=0.9\textwidth]{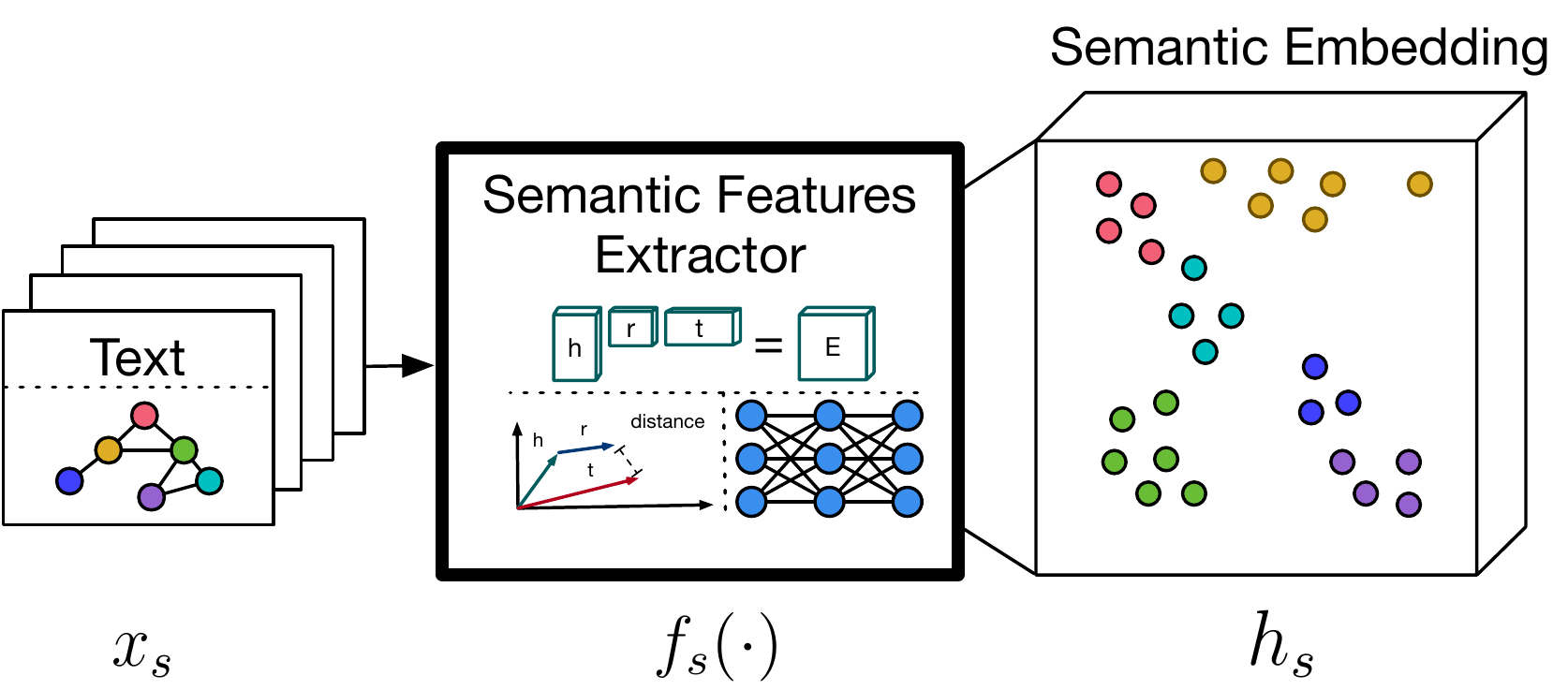}
    \caption{Semantic features extractor}
    \label{fig:semantic features extractor}
    \end{subfigure}
    \caption{Feature extractors transform input data into embedding space: a) a visual features extractor transforms visual input data, i.e. images, into visual embedding space; and b) a semantic features extractor transforms semantic input data, e.g. text or graphs, into semantic embedding space.}
    \label{fig:Feature extractor}
\end{figure*}

\paragraph{Visual Features Extractor:}
\label{sssec:visual features extractor}
A visual features extractor $f_v(\cdot)$, shown in Figure~\ref{fig:visual features extractor}, is a transformation function that transform visual input data $\vec{x}_v$ from an higher dimensional image space into a lower dimensional visual embedding space $\vec{h}_v$.

A formal definition is given by
\begin{equation}
\vec{h}_v = f_v(\vec{x_v}),
\label{eq:visual features extractor}
\end{equation}
where the final dimensionality of $\vec{h}_v$ is determined by the architecture.

Whereas early approaches used traditional visual features extractors as \emph{scale-invariant feature transform} (SIFT)\cite{DBLP:journals/ijcv/Lowe04} or \emph{histogram of oriented gradients} (HOG)~\cite{DBLP:conf/cvpr/DalalT05}, modern CV methods use almost only DNN-based approaches.
A common method to obtain a general DNN-based visual feature extractor is to pre-train a DNN on a large image dataset, such that the DNN automatically learns to extract valuable features out of the images.

\paragraph{Semantic Features Extractor:}
\label{sssec:semantic features extractor}
A semantic features extractor $f_s(\cdot)$, shown in Figure~\ref{fig:semantic features extractor}, is a transformation function that transform semantic input data $\vec{x}_s$ from an higher dimensional image space into a lower dimensional semantic embedding space $\vec{h}_s$.

A formal definition is given by
\begin{equation}
\vec{h}_s = f_s(\vec{x_s}),
\label{eq:visual features extractor}
\end{equation}
where the final dimensionality of $\vec{h}_s$ is determined by the architecture.

The term semantic data is here used for both, unstructured data from language and structured data from a KG.
Although the input data structure differs in its original format, the output of the semantic features extractor is always a low dimensional and vector-based semantic embedding space.
This similarity enables a seamless transfer from hybrid approaches of vision and language to hybrid approaches of vision and KGs.

\subsection{Knowledge Graph Embedding} 
\label{ssec:Knowledge Graph Embedding}
A knowledge graph embedding $h_{s}$ is a representation of a KG in vector space, where close relationships between entities in a KG are reflected by local neighborhoods in $h_{s}$.
$h_{s}$ is generated by a \emph{knowledge graph embedding method} (KGE-Method), which maps the entities and relations of a KG into low-dimensional vectors, while capturing their semantic meanings and relations~\cite{DBLP:journals/tkde/WangMWG17}.
Therefore, a KGE-Method is a special case of the semantic features extractors $f_s(\cdot)$ that works on graph data.

\begin{figure}[h]
\includegraphics[width=0.4\textwidth]{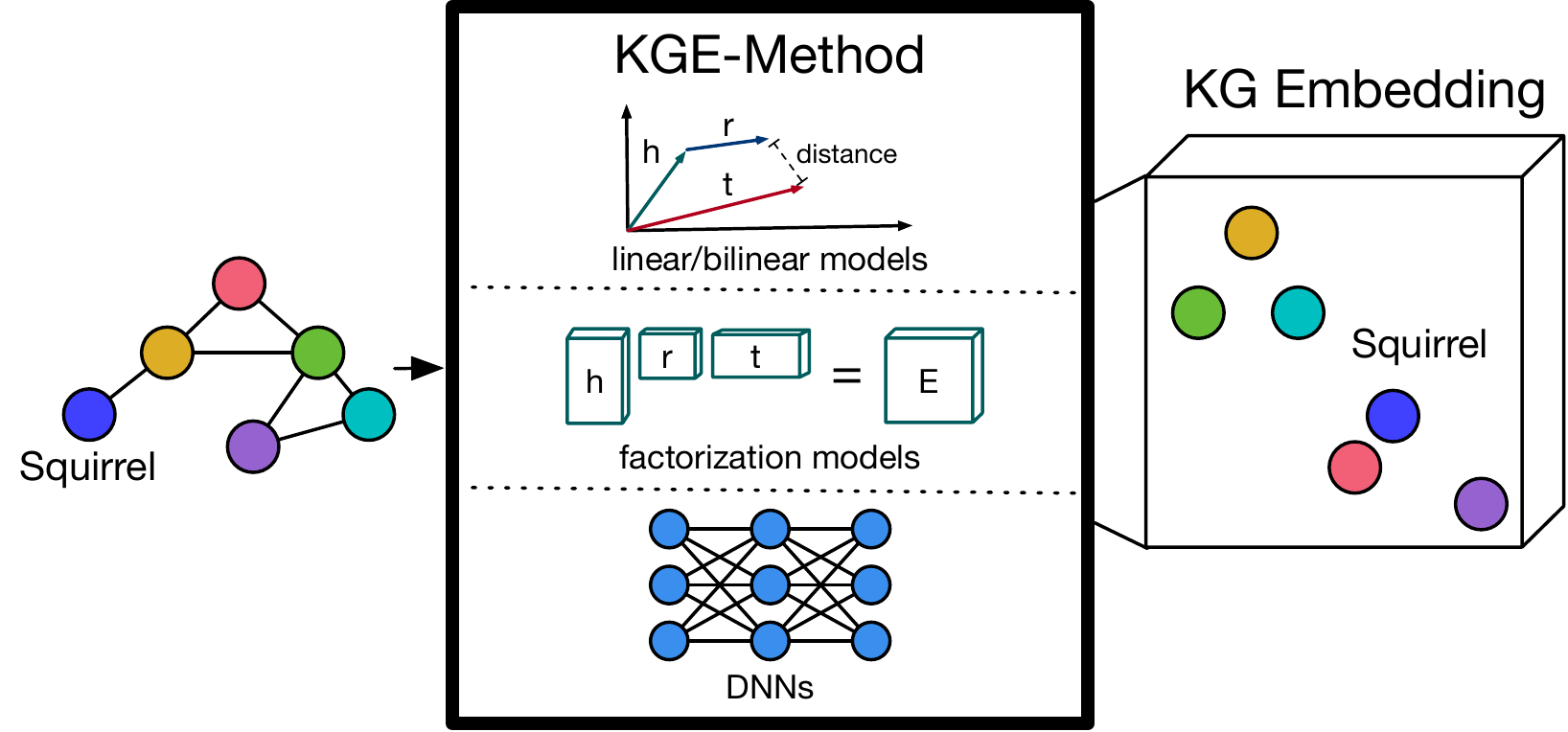}
\caption{A KGE-method transforms a KG into a knowledge graph embedding $h_s$.}
\label{fig:Knowledge graph embedding}
\end{figure}

In Figure~\ref{fig:Knowledge graph embedding}, the general pipeline of KGE-Methods which transform a KG into $h_s$ is illustrated.


\subsubsection{KGE-Methods - Learning Mode}
Originally, KGE-Methods were developed to solve graph-based tasks such as node classification or link prediction.
However, there is an increasing interest to apply KGE-Methods for visual tasks, such as classification, detection, or segmentation.
We briefly categorize KGE-Methods therefore into unsupervised and supervised KGE-Methods, as Chami et al.~\cite{DBLP:journals/corr/abs-2005-03675} recently proposed for graph embedding algorithms.

\paragraph{Unsupervised KGE-Methods:}
Unsupervised KGE-Methods form $h_{s}$ based on the inherent graph structure and the node features, without considering additional task-specific labels for the graph or its nodes.
An overview about unsupervised KGE-Methods is given by Ji et al.~\cite{DBLP:journals/corr/abs-2002-00388}, who categorized KGE-Methods based on their \emph{representation space} (vector, matrix, and tensor space), the \emph{scoring function} (distance-based, similarity-based), the \emph{encoding model} (linear/bilinear models, factorization models, neural networks), and the \emph{auxiliary information} (text descriptions, type constraints).

\paragraph{Supervised KGE-Methods:}
In contrast, supervised KGE-Methods learn $h_{s}$ to best predict node or graph labels.
Forming $h_{s}$ by using task-specific labels for the node features, $h_{s}$ can be optimized for a particular task while retaining the full expressivity of the graph.
The most common supervised KGE-Methods are \emph{graph neural networks} (GNNs)~\cite{1555942}. GNNs are extensions of standard DNNs that can directly work on a graph structure as provided by a KG.
For scalability reasons and to overcome challenges that arise from graph irregularities various adaptations have emerged, such as \emph{graph convolutional networks} (GCN)~\cite{DBLP:conf/iclr/KipfW17} or \emph{graph attention networks} (GAT)~\cite{DBLP:conf/iclr/VelickovicCCRLB18}.
Furthermore, non-Euclidean graph convolutional methods, such as \emph{hyperbolic graph convolutional neural networks} (HGCN)~\cite{DBLP:conf/nips/ChamiYRL19} are used to deal with a hierarchical structure of the input data.

\subsubsection{KGE-Methods - Input Type}
The majority of existing KGE-Methods only work on directed labeled graphs, expecting binary relations in a tripled-based format.
However, as shown in Section~\ref{ssec:Knowledge Graph}, a basic triplet representation oversimplifies the complex nature of the information that can be stored in hypergraphs and hyper-relational graphs~\cite{DBLP:conf/www/RossoYC20}.
A hypergraph or hypher-relational graph can be transformed into directed labeled graphs, either by \emph{reification}~\cite{DBLP:conf/ijcai/FatemiTV020}, that converts the graphs into binary-relation graphs, by creating additional triplets from a hyper-relational fact or by the \emph{star-to-clique}~\cite{DBLP:conf/ijcai/WenLMCZ16} technique, that converts a tuple defined on k entities into $k \choose 2$ tuples.
However, these conversions lead to suboptimal and incomplete models as well as information loss.
They only convert a set of key-value pairs, that are unaware of the triplet structure~\cite{DBLP:conf/www/RossoYC20, DBLP:conf/ijcai/FatemiTV020}.
To preserve the whole expressivity of the KG, a set of new KGE-Methods are developed to directly operate on hypergraphs and hyper-relational graphs.
Some of the methods that deal with hypergraphs are
HEBE~\cite{DBLP:conf/icdm/GuiLTJNH16}, HGE~\cite{DBLP:conf/cikm/YuTCY18}, Hyper2vec~\cite{DBLP:conf/dasfaa/HuangCYWZL19}, HNN~\cite{DBLP:conf/aaai/FengYZJG19}, HCN~\cite{DBLP:conf/nips/YadatiNYNLT19}, DHNE~\cite{DBLP:conf/aaai/TuCWW018}, HHNE~\cite{DBLP:conf/icdm/BaytasXW0Z18}, Hyper-SAGNN~\cite{DBLP:conf/iclr/ZhangZ020}, HypE~\cite{DBLP:conf/ijcai/FatemiTV020} and methods that embedd hyper-relational graphs are for instance
m-TransH~\cite{DBLP:conf/ijcai/WenLMCZ16}, HSimple~\cite{DBLP:conf/ijcai/FatemiTV020}, RAE~\cite{DBLP:conf/www/ZhangLMM18}, 
GETD\cite{DBLP:conf/www/0016Y020}, TuckER~\cite{DBLP:conf/emnlp/BalazevicAH19}, NaLP\cite{DBLP:conf/www/GuanJWC19}, 
HINGE\cite{DBLP:conf/www/RossoYC20}, StarE~\cite{DBLP:conf/emnlp/GalkinTMUL20}.

\subsection{Training Objectives for Joint Embeddings}
\label{sssec:Training objectives}
Since visual and semantic information can be encoded in a vector-based embedding space forming $h_v$ and $h_s$, there are several training objectives to learn a joint embedding.
%
The objectives and also the DNNs are optimized mainly using \emph{stochastic gradient descent} (SGD) or its derivatives.
SGD minimizes an objective, that measures how far apart the ground truth from the predicted probability distribution or value is.
The most common principle to derive specific objectives that are good estimators for different models is the maximum likelihood principle.
Any of these objectives can be seen as a cross entropy between the empirical distribution defined by the training set and the probability distribution defined by model~\cite{DBLP:books/daglib/0040158}.
Here we present some of the basic objectives used in visual transfer learning using KG, which can be augmented with additional regularization terms or hyperparameters.
Although work~\cite{DBLP:conf/eccv/BoudiafRZGPPA20,DBLP:journals/corr/abs-2010-16402} showed that the objectives have a smaller impact on the learned DNN than suspected, there are configurations of visual and semantic embedding space that only allow certain objectives to be applied.
We define $\vec{l} \in \mathbb{R}^K$ as the network’s output (logits) vector, and $\vec{t} \in {0, 1}^K$ as the one-hot encoded vector of targets, where $\left\lVert{t}\right\rVert_1 = 1$.
We refer to visual data as $x_v$ and semantic data as $x_s$, and equally to visual embedding as $h_v$ and semantic embedding as $h_s$.

\subsubsection{Pointwise Objectives}

\paragraph{Softmax Cross-Entropy (CE)~\cite{DBLP:conf/nato/Bridle89}:}
CE is the most common objective to learn multi-class classification tasks.
The softmax represents a probability distribution over a discrete variable with $K$ possible values, i.e. classes.
CE learns the DNN end-to-end by comparing the logits $\vec{l}$ with the target vector $\vec{t}$ and is given by
\begin{align}
L_{CE}(\vec{l},\vec{t}) & = - \sum_{k=1}^{K}{t_k \log{\left(\frac{\exp{(l_k)}}{\sum_{j=1}^{K}{\exp{(l_j)}}}\right)}} \\
                             & = - \sum_{k=1}^{K}{t_k l_k} + \log{\sum_{k=1}^{K} \exp{(l_k)}}
\label{eq:softmax}
\end{align}

\paragraph{Mean Squared Error (MSE):}
MSE is the most intuitive way of attracting two vectors and is given by
\begin{equation}
L_{MSE} = \frac{1}{K}\sum_{k = 1}^K \left\|\vec{h}_{s,k} - \vec{h}_{v,k})\right\|^2.
\label{eq:mse}
\end{equation}
The MSE loss calculates the Euclidean distance and maps a training image $x_{v,k}$ and its visual feature vector $h_{v,k)}$ to a semantic embedding vector $h_{s,k}$, corresponding to the same class $k$~\cite{DBLP:conf/nips/SocherGMN13}.

However, using the Euclidian distance as a metric fails in high-dimensional space~\cite{Maaten2008VisualizingDU}.
An alternative metric in high dimensions is the cosine distance, which is given by
$sim(\vec{u}, \vec{v}) = \vec{u}^\top\vec{v} / \left\|\vec{u}\right\| \left\|\vec{v}\right\|$.

\subsubsection{Pairwise Objectives}
Pairwise objectives~\cite{DBLP:conf/cvpr/HadsellCL06} always rely on the information of positive and negative samples.
They have the goal to pull positive visual embedding vectors $\vec{h}_{v,p}$ to its corresponding semantic embedding anchor vector $\vec{h}_{s,a}$ and push negatives $\vec{h}_{v,n}$ away~\cite{DBLP:conf/nips/FromeCSBDRM13}.

\paragraph{Triplet and Hinge Rank Loss~\cite{DBLP:conf/cvpr/WangSLRWPCW14}:}
The triplet and hinge rank loss requires an explicit negative sampling.
It uses a margin $\alpha$ as a regularization term and it is given by
\begin{equation}
\hspace*{-0.6cm}
L_{tri} = \sum_{n \neq p} max[0, \alpha - sim(\vec{h}_{s,a}, \vec{h}_{v,p}) + sim(\vec{h}_{s,a}), \vec{h}_{v,n}].
\label{eq:hinge-rank}
\end{equation}

\paragraph{Contrastive Loss:}
The contrastive loss extends the triplet loss by a version of the softmax and handles multiple positives and negatives at a time and is given by
\begin{equation}
\hspace*{-0.6cm}
L_{con} = - \log \frac{\exp{(sim(\vec{h}_{s,a},\vec{h}_{v,p}) / \tau)}}{\sum^{2N}_{n=1}\mathds{1}_{n \neq a}\exp{(sim(\vec{h}_{s,a}, \vec{h}_{v,n}) / \tau)}}
\label{eq:2}
\end{equation}
where, $\mathds{1}_{n \neq a} \in \{0, 1\}$ is an indicator function that returns 1 iff $n \neq a$, and $\tau > 0$ denotes a temperature parameter.

\section{Visual Transfer Learning using Knowledge Graphs}
\label{sec:Visual Transfer Learning using Knowledge Graphs}

Visual transfer learning using knowledge graphs has proven to be particularly advantageous compared to approaches without auxiliary knowledge~\cite{DBLP:conf/nips/SocherGMN13, DBLP:conf/cvpr/0004YG18}.
Since auxiliary knowledge mitigates the sole dependence on data distribution, it leads to models that are better generalized and thus more robust and applicable to new domains~\cite{DBLP:conf/aaai/LarochelleEB08}.
Having various kinds of auxiliary knowledge, a KG can serve as a universal knowledge representation.
KGs encode the classes either hierarchically, organized in superclasses, or flat, using relationships to other objects or other classes.
Section~\ref{ssec:Knowledge Graph} presents three distinct modeling structures with different levels of expressiveness and Section~\ref{ssec:Knowledge Graph Embedding} introduces relevant embedding methods.
All approaches that use a KG in combination with a DNN use the KG to implement some prior assumptions in the data-driven DL pipeline.
A prior assumption induced by the KG is the definition of relationships between objects/classes so that objects/classes can borrow statistical strength from other related objects/classes in the graph.
These priors give the CV process a structure that allows making better predictions even when visual data is sparse or erroneous.
However, there are several ways the auxiliary knowledge of a KG can be induced into a DNN.

Referring to \textbf{RQ1}, this section provides a categorization of visual transfer learning approaches that combine KGs with the DL pipeline.

\begin{figure*}[htb]
\includegraphics[width=\textwidth]{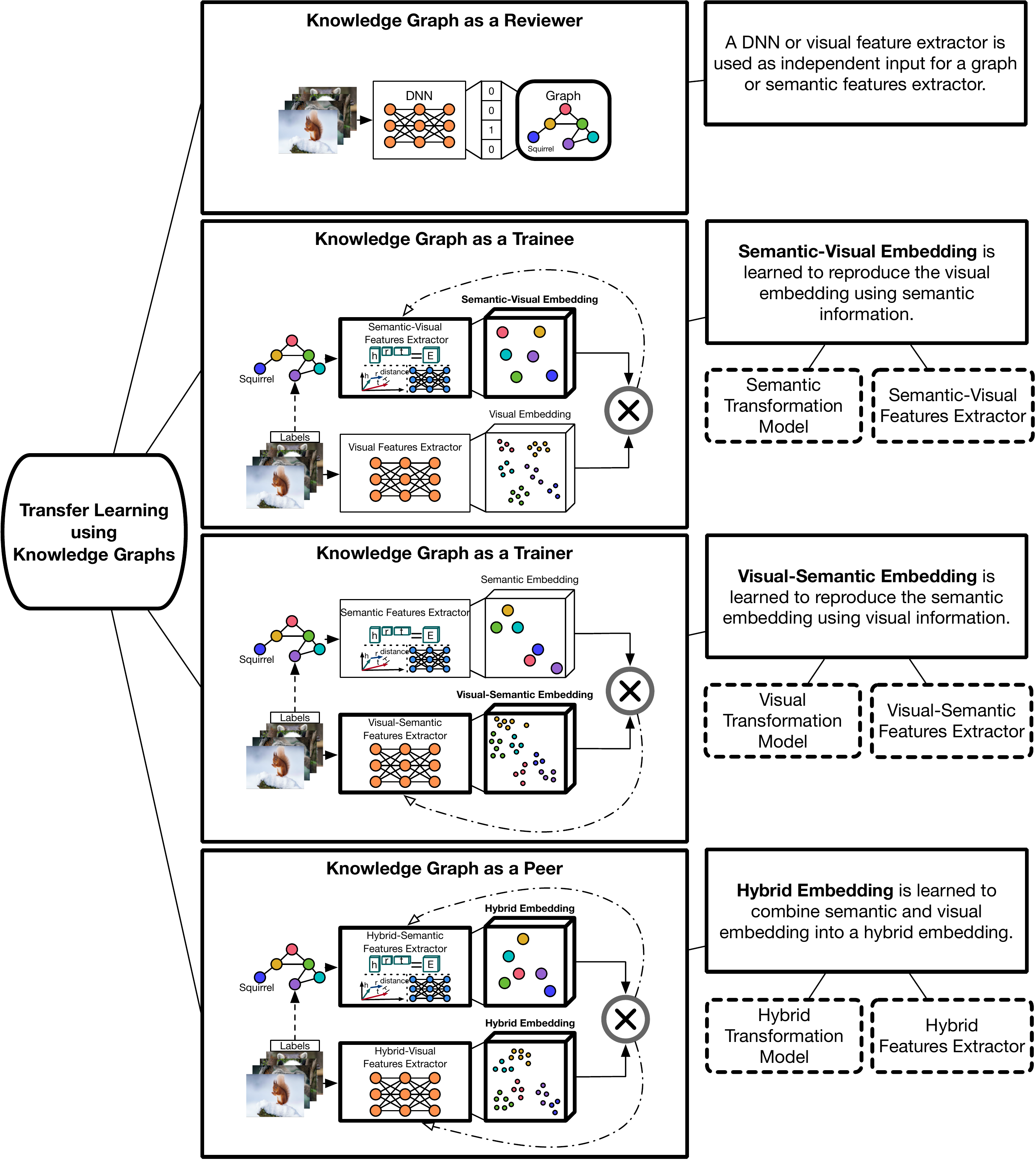}
\caption{Visual transfer learning using KGs according to the role of the KG are split in four categories: 1) \emph{Knowledge Graph as a Reviewer}; 2) \emph{Knowledge Graph as a Trainee}; 3) \emph{Knowledge Graph as a Trainer}; and 4) \emph{Knowledge Graph as a Peer}.}
\label{fig:categories}
\end{figure*}

As shown in Figure~\ref{fig:categories}, we categorize the field of visual transfer learning using knowledge graphs into:\\
1) \emph{Knowledge Graph as a Reviewer} - where the KG is used for post-validation of a visual model;\\
2) \emph{Knowledge Graph as a Trainee}, where a semantic-visual embedding $h_{s,v}$ is learned using a visual embedding $h_v$ as objective;\\
3) \emph{Knowledge Graph as a Trainer}, a visual-semantic embedding $h_{v,s}$ is learned using a semantic embedding $h_s$ as objective; and\\
4) \emph{Knowledge Graph as a Peer}, where a hybrid-embedding $h_h$ is learned using a combination of semantic embedding $h_s$ and a visual embedding $h_v$ as objective.\\
Since KGE-Methods have only recently entered the field of visual transfer learning, we also list related methods forming $h_s$ based on other types of auxiliary knowledge in categories 2), 3), and 4).
Other types of auxiliary knowledge are language descriptions or class attributes, so that their semantic features extractor $f_{s}(\cdot)$ differs in the type of input, but not in its architecture, as described in Section~\ref{sssec:semantic features extractor}.

Regarding \textbf{RQ2}, we describe the categories and their approaches in detail and discuss their field of application and their properties.
A summary of all approaches and their respective transfer learning task is given in Table~\ref{tab:categories_and_their_tasks}.

\subsection{Knowledge Graph as a Reviewer}
\label{ssec:Knowledge Graph as a Reviewer}

\begin{figure}[h]
\includegraphics[width=0.45\textwidth]{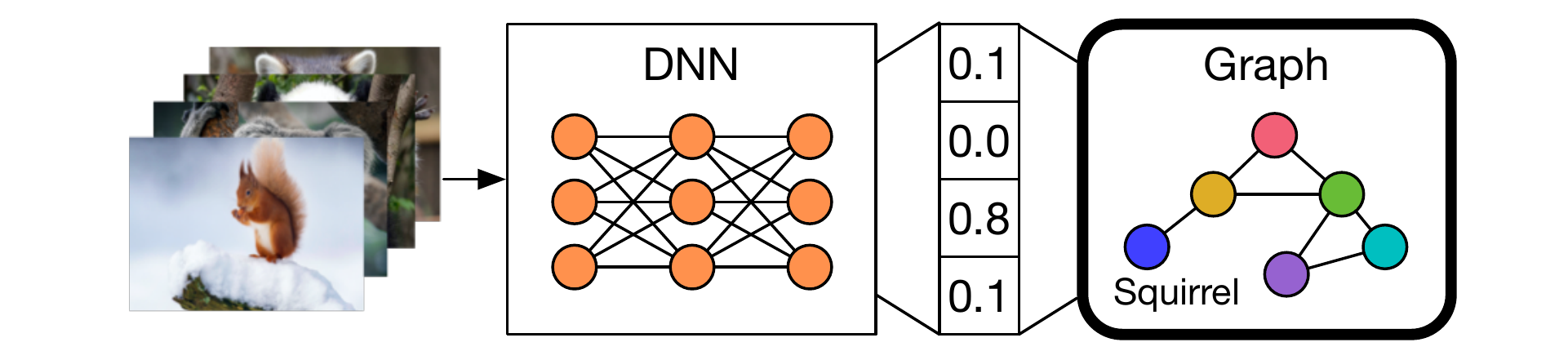}
\caption{Approaches from the category \emph{Knowledge Graph as a Reviewer} use the KG for post-validation of a pre-trained DNN or its intermediate feature layers.}
\label{fig:knowledge graph as a reviewer}
\end{figure}

Approaches of the category \emph{Knowledge Graph as a Reviewer} arrange the visual model and the KG in a sequential order, as depicted in Figure~\ref{fig:knowledge graph as a reviewer}.
The visual output of a pre-trained DNN or its intermediate feature layers suit as an input to a graph or graph-based network.
Unlike the other categories, the KG as a reviewer does not learn a joint embedding space, instead, it uses the KG or its $h_s$ to reason over the independent output of a visual model $h_v$.

Most of the approaches map the output of a visual features extractor $f_{v}(\cdot)$ on the corresponding input nodes in a hierarchical graph, to enrich the output with inter-class relationships.
Lampert et al.~\cite{DBLP:conf/cvpr/LampertNH09} train a \emph{support vector machine} (SVM) on SIFT features to predict binary \emph{animals with attributes} (AwA) dataset attributes. 
These class attributes are fed into a hierarchical graph-based network to predict unknown classes for a zero-shot learning task.
Salakhutdinov et al.~\cite{DBLP:conf/cvpr/SalakhutdinovTT11} introduce a hierarchical Bayesian classification model~\cite{Shahbaba2005ImprovingCW} that learns a tree structure of class and super-class relationships.
They use their learned graph on top of an SVM, which classifies HOG features of images.
They show that their method using a learned graph outperforms a method using a fixed graph based on WordNet\footnote{\url{https://wordnet.princeton.edu/}}~\cite{DBLP:journals/cacm/Miller95} and other approaches without hierarchical graph information.
Deng et al.~\cite{DBLP:conf/cvpr/DengKBF12} proposed the \emph{DARTS} algorithm for zero-shot learning.
They pre-train an SVM on SIFT features of the ImageNet~\cite{DBLP:conf/cvpr/DengDSLL009} dataset and map its classification output to WordNet with a reward and an accuracy to maximize the information gain.
Ordonez et al.~\cite{DBLP:conf/iccv/OrdonezDCBB13} extend the approach to output human-understandable entry categories for images.
They enrich the output of an SVM-based image classification model with information from a text-based n-gram language model by mapping both sources to the corresponding node in the WordNet graph.
Rohrbach et al.~\cite{DBLP:conf/nips/RohrbachES13} present \emph{propagated semantic transfer} (PST).
They use WordNet and attribute vectors from the AwA dataset to perform classification on few-shot learning classes of ImageNet.
PST exploits similarities in visual embeddings of known classes encoded by an SVM learning a \emph{k-Nearest Neighbor} (kNN) graph that helps to find relationships to new classes.
Deng et al.~\cite{DBLP:conf/eccv/DengDJFMBLNA14} propose to use a \emph{hierarchy and exclusion} (HEX) graph that exploits hierarchical class relationships of the output of a visual model.
HEX graphs allow flexible specification of relations between labels applied to the same object.
To build the graph, they use the hierarchical structure of WordNet extended with additional specifications and relations to objects, such as mutual exclusion (e.g., an object cannot be a dog and a cat), overlap (e.g., a husky can be a puppy and vice versa), and subsumption (e.g., all huskies are dogs).
In addition, they proposed a probabilistic classification model that exploits their HEX graphs and evaluated their approach on ImageNet, in object classification and zero-shot learning.
Gebru et al.~\cite{DBLP:conf/iccv/GebruHF17} use WordNet attributes to improve fine-grained object classification on the task of domain generalization with the Office-31~\cite{DBLP:conf/eccv/SaenkoKFD10} and the large-scale Car dataset~\cite{DBLP:conf/aaai/GebruKWCDF17}.
Source and target domain images are fed through a pipeline with two identical CNNs and a classification layer that classifies both the fine-grained classes and the different attribute types.
The Kullback–Leibler divergence is used to compare the predicted label distributions.
Lee et al.~\cite{DBLP:conf/cvpr/LeeFYW18} propose a \emph{graph gated neural network} (GGNN) that incorporates a structured KG based on WordNet and learned edge weights to improve zero-shot learning.
First, an NN is learned that combines the GloVe~\cite{DBLP:conf/emnlp/PenningtonSM14} language embeddings of the class labels and the pre-trained visual feature vectors of the images as input to the GGNN.
Second, the GGNN learns to propagate the information through the KG and outputs a final probability for each node.


\begin{table*}[htb]
    \centering
    \begin{tabular}{|p{3.9cm} | p{2.6cm} | p{2.5cm} | p{2.5cm} | p{2.5cm}|}
        \hline
        \textbf{Category} & \textbf{Sub-Category} & \textbf{Task Transfer} & \textbf{Domain Transfer} & \textbf{Other} \\
        \hline
        Knowledge Graph as a Reviewer
        & 
        & \cite{DBLP:conf/cvpr/LampertNH09},
        \cite{DBLP:conf/cvpr/DengKBF12},
        \cite{DBLP:conf/nips/RohrbachES13},
        \cite{DBLP:conf/eccv/DengDJFMBLNA14},
        \cite{DBLP:conf/aaai/LuoZHY20}
        & \cite{DBLP:conf/iccv/GebruHF17},
        \cite{DBLP:conf/cvpr/Gong0LS0L19}
        & \cite{DBLP:conf/cvpr/SalakhutdinovTT11},
        \cite{DBLP:conf/iccv/OrdonezDCBB13},
        \cite{DBLP:conf/cvpr/MarinoSG17},
        \cite{DBLP:conf/cvpr/ChenLFG18},
        \cite{DBLP:conf/nips/JiangXLL18},
        \cite{DBLP:journals/corr/abs-1908-04385},
        \cite{DBLP:conf/nips/LiangHZLX18}
        \\
        \hline
        \multirow{2}{*}{Knowledge Graph as a Trainee}
        & Semantic-Visual\newline Transformation Model 
        & \textcolor{Red}{\cite{DBLP:conf/cvpr/RochanW15}},
        \textcolor{Red}{\cite{DBLP:conf/cvpr/ZhangXG17}}
        & 
        & \\
        \cline{2-5}
        & Semantic-Visual\newline Features Extractor 
        & \cite{DBLP:conf/cvpr/0004YG18},
        \cite{DBLP:conf/cvpr/KampffmeyerCLWZ19},
        \cite{DBLP:conf/aaai/GaoZX19},
        \cite{DBLP:conf/iccv/PengLZLQT19},
        \cite{DBLP:conf/aaai/ChenCHWLL20},
        \cite{DBLP:conf/www/GengC0PYYJC21},
        \textcolor{Red}{\cite{DBLP:conf/cvpr/ZhuEL0E18}},
        \textcolor{Red}{\cite{DBLP:conf/cvpr/LiJLD0H19}},
        \textcolor{Red}{\cite{DBLP:conf/eccv/VyasVP20}}
        & 
        & \cite{DBLP:conf/cvpr/ChenWWG19}
        
        \\
        \hline
        \multirow{2}{*}{Knowledge Graph as a Trainer}
        & Visual-Semantic\newline Transformation Model 
        & 
        \cite{DBLP:journals/pami/AkataPHS16},
        \textcolor{Red}{\cite{DBLP:conf/nips/PalatucciPHM09}},
        \textcolor{Red}{\cite{DBLP:conf/nips/SocherGMN13}},
        \textcolor{Red}{\cite{DBLP:conf/nips/FromeCSBDRM13}},
        \textcolor{Red}{\cite{DBLP:journals/corr/NorouziMBSSFCD13}},
        \textcolor{Red}{\cite{DBLP:conf/iccv/ZhangS15a}},
        \textcolor{Red}{\cite{DBLP:conf/cvpr/KodirovXG17}}
        & 
        & \textcolor{Red}{\cite{Mitchell1191}}
        \\
        \cline{2-5}
        & Visual-Semantic\newline Features Extractor 
        & \cite{DBLP:conf/pkdd/JayathilakaMS21}
        & \cite{DBLP:conf/semweb/MonkaH0R21},
        \textcolor{Red}{\cite{radford2learning}}
        & \textcolor{Red}{\cite{DBLP:conf/eccv/JoulinMJV16}}
        
        \\        
        \hline
        \multirow{2}{*}{Knowledge Graph as a Peer} 
        & Hybrid\newline Transformation Model 
        & 
        \cite{DBLP:conf/iccv/ZhaoPZF017},
        \cite{DBLP:journals/corr/YangH14a},
        \cite{DBLP:journals/corr/abs-2012-06236},
        \textcolor{Red}{\cite{DBLP:journals/pami/FuHXG15}},
        \textcolor{Red}{\cite{DBLP:conf/iccv/BaSFS15}},
        \textcolor{Red}{\cite{DBLP:conf/cvpr/ChangpinyoCGS16}},
        \textcolor{Red}{\cite{DBLP:conf/iccv/TsaiHS17}},
        \textcolor{Red}{\cite{DBLP:conf/iccv/Jiang0SC19}}
        & \textcolor{Red}{\cite{DBLP:journals/corr/YangH14a}}
        & \textcolor{Red}{\cite{DBLP:conf/cvpr/KarpathyL15}},
        \textcolor{Red}{\cite{DBLP:journals/pami/TangWWGDGC18}},
        \textcolor{Red}{\cite{DBLP:conf/eccv/Li0LZHZWH0WCG20}},
        \textcolor{Red}{\cite{DBLP:journals/corr/abs-2006-16934}}
        \\
        \cline{2-5}
        & Hybrid\newline Features Extractor 
        & \cite{DBLP:journals/corr/abs-2102-01987}
        & 
        & \textcolor{Red}{\cite{DBLP:journals/corr/abs-2010-00747}}
        \\        
        \hline
    \end{tabular}
    \caption{Categories and their tasks: Task transfer refers to the category zero and few-shot learning, domain transfer refers to the category domain generalization and adaptation, and other relates to object classification, object detection, and object segmentation on source task and domain only. Note: All approaches using related types of auxiliary knowledge are highlighted in red.}
    \label{tab:categories_and_their_tasks}
\end{table*}

Instead of using hierarchical graphs of WordNet and class attributes only, other approaches make use of flat object or class relationships.
Their graph consists of specific real-world configurations of objects and their appearance.
Marino et al.~\cite{DBLP:conf/cvpr/MarinoSG17} improves fine-grained image classification by creating a KG using the most common object-attribute and object-object relationships of the Visual Genome~\cite{DBLP:journals/ijcv/KrishnaZGJHKCKL17} dataset and higher-level semantics from WordNet.
The output of a pre-trained, faster R-CNN~\cite{DBLP:journals/pami/RenHG017} object detector is fed into a \emph{graph search neural network} (GSNN) which reasons about relationships of the detected objects.
The final prediction is a combination of the GSNN output, the visual embedding, and the detections of the faster R-CNN.
Chen et al.~\cite{DBLP:conf/cvpr/ChenLFG18} propose an object detection post-processing that connects a local and a global module via an attention mechanism.
The local module is based on a convolutional \emph{gated recurrent unit} (GRU) and builds spatial memory of previously detected objects using the class label and its visual embedding.
The global graph-reasoning module consists of two paths, a spatial path that uses a region graph to connect far detected classes, and a semantic path which uses a KG, based on ADE20K~\cite{DBLP:journals/ijcv/ZhouZPXFBT19} and Visual Genome, to connect classes with semantically related classes.
Jiang et al.~\cite{DBLP:conf/nips/JiangXLL18} extend~\cite{DBLP:conf/cvpr/ChenLFG18} with \emph{hybrid knowledge routed modules} (HKRM) allowing them to be applied on the intermediate feature representation directly to check the compatibility of auxiliary knowledge with visual evidence in each image.
HKRM can be divided into an explicit knowledge module and an implicit knowledge module, whereas the former contains external knowledge such as shared attributes, co-occurrence, and relationships, and the latter is built without explicit definitions and forms a region-to-region graph with constraints over objects, as spatial knowledge such as layout, size, overlap.
Liu et al.~\cite{DBLP:journals/corr/abs-1908-04385} improve object detection by feeding the final object detections into a GCN which is based on object relationships and learned from MSCOCO dataset~\cite{DBLP:conf/eccv/LinMBHPRDZ14}.
Gong et al.~\cite{DBLP:conf/cvpr/Gong0LS0L19} propose a human parsing agent called "Graphonomy" that learns a knowledge graph on a conventional parsing network.
It consists of an intra-graph reasoning module in form of a GCN whose structure uses semantic constraints from the human body to transfer knowledge within a dataset due to encoded relationships between nodes, and an inter-graph reasoning module, that uses handcrafted relations, a learnable matrix, feature similarities, and semantic similarities, to transfer semantic information between different datasets.
Liang et al.~\cite{DBLP:conf/nips/LiangHZLX18} present a \emph{symbolic graph reasoning} (SGR) layer for semantic segmentation and image classification.
It consists of a module that assigns the visual features of a pre-trained DNN to corresponding nodes of a KG.
A graph reasoning over all previously defined nodes is performed, and a mapping from the symbolic graph information back to the visual feature space.
Their graph is based on an object relation graph from Visual Genome and a hierarchical relation graph from WordNet.

Luo et al.~\cite{DBLP:conf/aaai/LuoZHY20} propose a context-aware zero-shot learning framework, where they use a KG to reason about visual feature vectors generated from an object detection model.
By using inter-class relationships, they improve traditional zero-shot learning techniques on the Visual Genome dataset.





\subsection{Knowledge Graph as a Trainee}
\label{ssec:Knowledge Graph as a Trainee}
\begin{figure*}[t]
    \centering
        \begin{subfigure}[t]{0.49\textwidth}
    \centering
    \includegraphics[width=\textwidth]{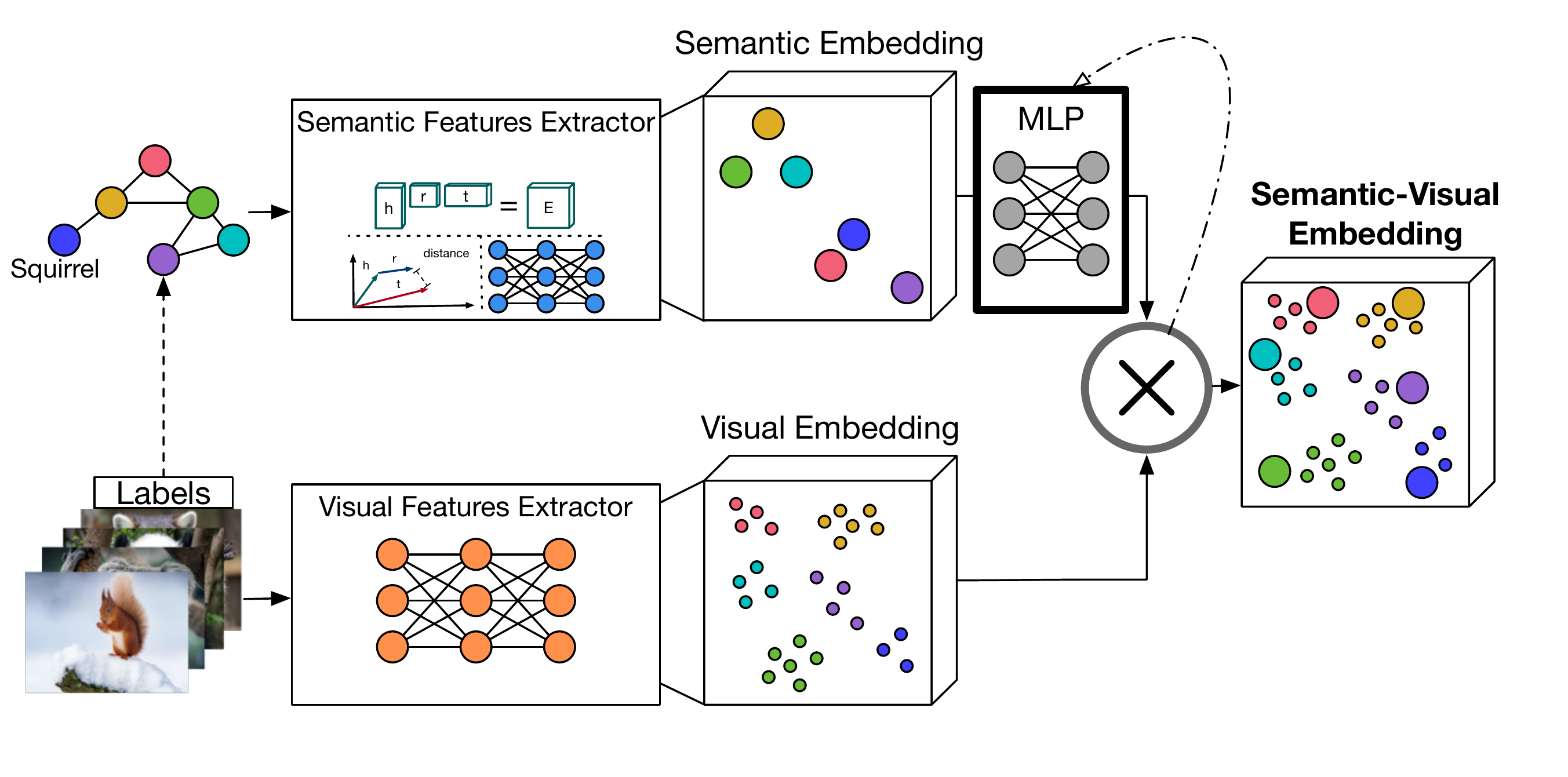}
    \caption{Semantic-Visual Transformation Model}
    \label{fig:Knowledge Graph as a Trainee MLP}
    \end{subfigure}
    ~
    \begin{subfigure}[t]{0.49\textwidth}
    \centering
    \includegraphics[width=\textwidth]{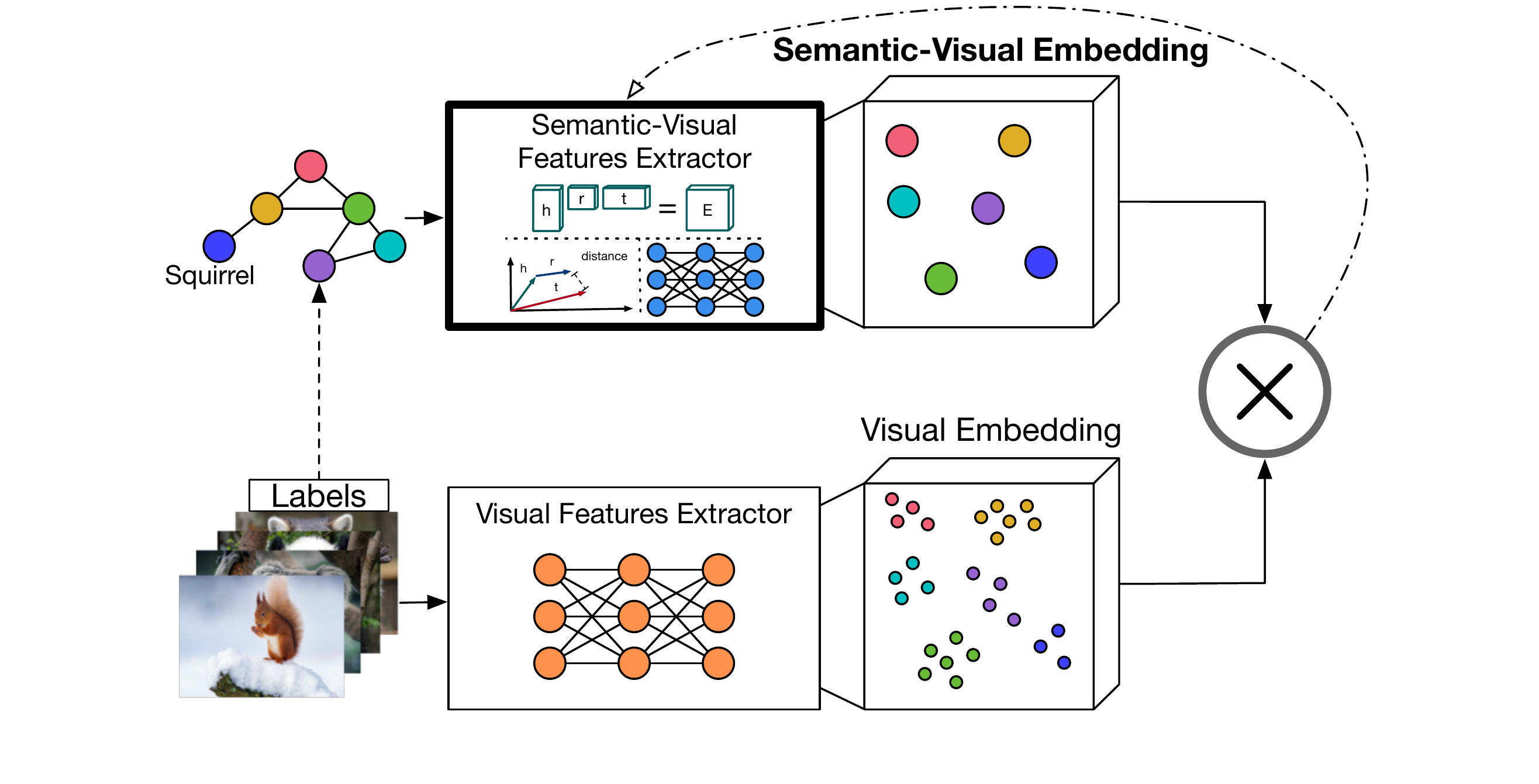}
    \caption{Semantic-visual features extractor}
    \label{fig:Knowledge Graph as a Trainee}
    \end{subfigure}
    \caption{Approaches that belong to the category \emph{Knowledge Graph as a Trainee} learn semantic visual embedding space supervised by a visual embedding. They either learn a) a transformation function, e.g. MLP, on top of a pre-trained semantic embedding space or b) a semantic-visual features extractor.}
    \label{fig:Knowledge Graph as a Trainee overview}
\end{figure*}

Approaches that belong to this category combine the visual DNN with the auxiliary knowledge of a KG by learning a semantic-visual embedding $h_{s,v}$.
Unlike the \emph{Knowledge Graph as a Reviewer}, which uses the visual embedding $h_v$ as input for the KG, approaches from the category \emph{Knowledge Graph as a Trainee} use $h_v$ as an objective to embedd the KG into $h_{s,v}$.
Figure~\ref{fig:Knowledge Graph as a Trainee overview} illustrates a conceptual architecture of the knowledge graph as a trainee approach.
To combine visual and semantic information, some approaches either learn a transformation function, e.g. MLP, on top of a semantic embedding space $h_s$, or apply supervised KGE-Methods to learn a semantic-visual features extractor $f_{s,v}(\cdot)$ directly.

\subsubsection{Semantic-Visual Transformation Models}
As shown in Figure~\ref{fig:Knowledge Graph as a Trainee MLP}, the pre-trained $h_{s}$ is fixed over the whole training process, and an additional transformation function, e.g. MLP, is learned to transform $h_{s}$, into the semantic-visual embedding space $h_{s,v}$.

\paragraph{Related Approaches using other Auxiliary Knowledge:}
Rochan et al.~\cite{DBLP:conf/cvpr/RochanW15} used a fixed language embedding to define relationships between classes, that unknown classes in a zero-shot learning task can borrow their visual embeddings from a linear combination of known related classes.
Zhang et al.~\cite{DBLP:conf/cvpr/ZhangXG17} extends suggesting to use the visual space, instead of the semantic space, as the main embedding space, thus reducing the hubness problem that occurs in high dimensions.





\subsubsection{Semantic-visual Features Extractors}

As illustrated in Figure~\ref{fig:Knowledge Graph as a Trainee} the semantic-visual features extractor $f_{s,v}(\cdot)$ learns to directly transform the KG into a semantic-visual embedding $h_{s,v}$ using the supervision of the visual embedding space $h_v$.
As described in Section~\ref{ssec:Knowledge Graph Embedding}, $f_{s,v}(\cdot)$ is mostly implemented using a supervised KGE-Method.

Wang et al~\cite{DBLP:conf/cvpr/0004YG18} build a GCN on the structure of WordNet and optimize it to predict ImageNet pre-trained visual classifiers.
Based on the learned relations in the GCN they are able to transform information to novel class nodes to perform zero-shot learning.
A similar principle is used by Chen et al.~\cite{DBLP:conf/cvpr/ChenWWG19} for multi-label image recognition.
However, instead of using a hierarchical graph, the approach uses an object-relation graph which reflects the different relations between objects in a scene.
They build their graph based on the occurrence probabilities of different objects in the MSCOCO dataset since some objects are more likely to occur together.
Kampffmeyer et al.~\cite{DBLP:conf/cvpr/KampffmeyerCLWZ19} claim that multi-layer GNN architectures, which are required to propagate knowledge to distant nodes in the graph, dilute the knowledge by performing extensive Laplacian smoothing at each layer and thereby consequently decrease performance.
They propose a \emph{dense graph propagation} (DGP) module with direct links among distant nodes to exploit the hierarchical graph structure of the KG.
They tested their approach on zero-shot learning tasks as 21K ImageNet dataset and AWA2.
Gao et al.~\cite{DBLP:conf/aaai/GaoZX19} designed a \emph{two-stream GCN} (TS-GCN) to perform \emph{zero-shot action recognition} (ZSAR).
Their GCN architectures are based on the ConceptNet 5.5 KG, which contains information from various knowledge bases such as WordNet and DBpedia.
The first classifier branch uses the language embedding vectors of all classes as input for a GCN and then generates the classifiers for each action category.
The second instance branch feeds video segments into a DNN and outputs object scores, which are combined with attribute vectors from the classifier branch using a post-processing GCN  to form an attribute feature space.
The final objective is then defined by a comparison of the attribute feature space and the output of the classifier branch.
Peng et al.~\cite{DBLP:conf/iccv/PengLZLQT19} propose a \emph{knowledge transfer network} (KTN), which extends~\cite{DBLP:conf/cvpr/0004YG18} with a vision-knowledge fusion model.
This vision-knowledge fusion model is used to combine the final prediction output of the GCN with the output of a DNN, as they claim that semantic embeddings and visual embeddings are complementary and therefore cannot be combined with a single inner product.
They pre-train their visual feature learning module using cosine similarity on image data, use a subgraph of WordNet for their knowledge transfer module, and language embeddings of the class labels as the initial state of the nodes of the GCN.
Chen et al.~\cite{DBLP:conf/aaai/ChenCHWLL20} present the \emph{knowledge graph transfer network} (KGTN).
The knowledge graph transfer module incorporates a GGNN, which supports knowledge transfer of classes through a KG.
To train GGNN, they fix the weights of a pre-trained visual features extractor and examine three different similarity metrics, such as inner product, cosine similarity, and person correlation coefficient, to compare the output of the DNN and the GGNN.
They show that the accuracy of the model benefits from a reasoning process and the auxiliary knowledge from a KG.

Geng et al.~\cite{DBLP:conf/www/GengC0PYYJC21} recently proposed Onto-ZSL, an ontology-enhanced zero-shot learning framework that can be applied either to image classification or knowledge graph completion.
They build an inter-class relationship using an ontological schema, that comprises a label taxonomy from WordNet, textual descriptions, and attribute descriptions.
Further, they address the data imbalance problem between seen and unseen images by leveraging a \emph{generative adversarial network} (GAN) that produces synthesized visual feature vectors for unseen classes.

\paragraph{Related Approaches using other Auxiliary Knowledge:}
Approaches using language models leverage GANs to imagine unseen categories from text descriptions and hence recognize novel classes with no examples being seen.
GANs can be seen as a transformation function from text-based input to visual features, using the supervision of a visual model.
Zhu et al.~\cite{DBLP:conf/cvpr/ZhuEL0E18} propose GAZL, an approach that takes noisy text descriptions about unseen classes from Wikipedia and generates synthesized visual features for this class.
Using textual input for unseen classes they learn a GAN that generates visual features similar to the pre-trained ones of the seen classes.
Therefore, the zero-shot learning problem is transformed into a standard classification task and a classifier that can handle unseen classes can be trained using the synthesized image features for every unseen class.
Li et al.~\cite{DBLP:conf/cvpr/LiJLD0H19} extended the approach by introducing LisGAN, a GAN that takes semantic descriptions and random noise to generate visual features for unseen classes.
In addition, they deploy the average representation of all samples from an unseen class defining the soul sample of the class to reduce the noise in the predictions.
Vyas et al.~\cite{DBLP:conf/eccv/VyasVP20} propose LsrGAN, a generative model that leverages the semantic relationship between seen and unseen categories and explicitly performs knowledge transfer by incorporating a novel \emph{semantic regularized loss} (SR-Loss).
Knowing the inter-class relationships in the semantic space helps to impose the same relationship constraints
among the generated visual features.




\subsection{Knowledge Graph as a Trainer}
\label{ssec:Knowledge Graph as a Trainer}
\begin{figure*}[t]
    \centering
    \begin{subfigure}[t]{0.49\textwidth}
    \centering
    \includegraphics[width=\textwidth]{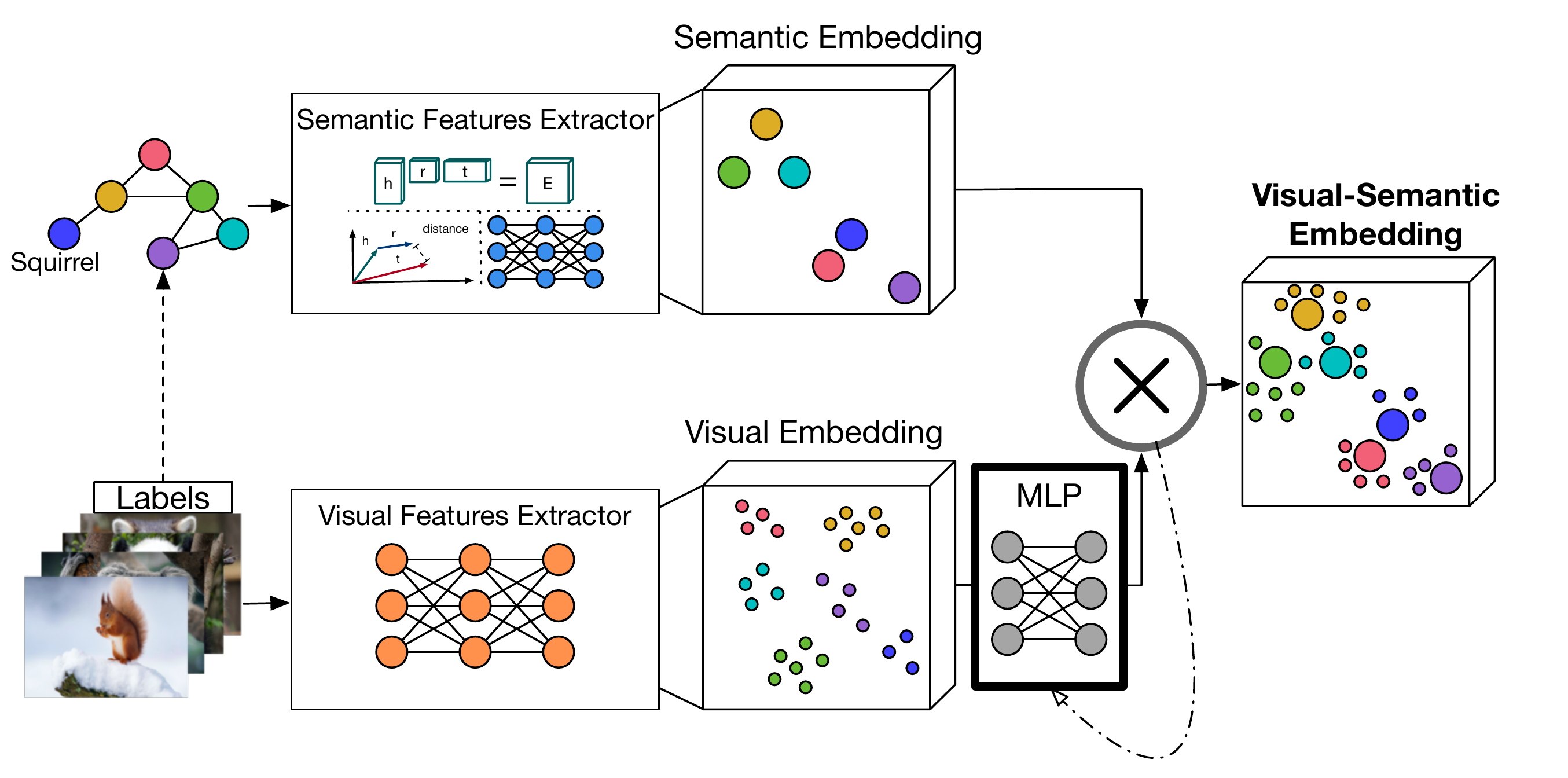}
    \caption{Visual-semantic transformation model}
    \label{fig:Knowledge Graph as a Trainer MLP}
    \end{subfigure}
    ~
    \begin{subfigure}[t]{0.49\textwidth}
    \centering
    \includegraphics[width=\textwidth]{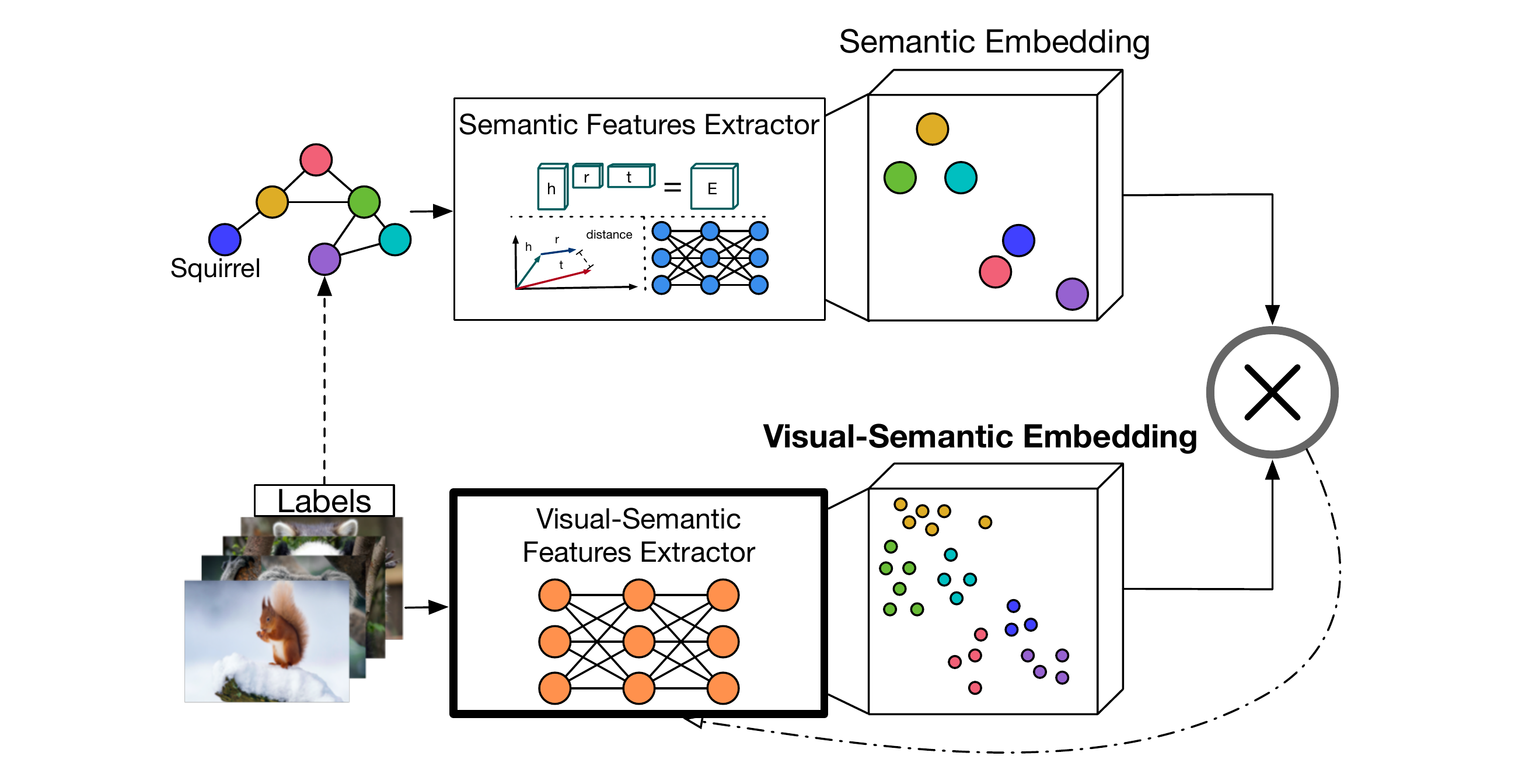}
    \caption{Visual-semantic features extractor}
    \label{fig:Knowledge Graph as a Trainer}
    \end{subfigure}
    \caption{Approaches that belong to the category \emph{Knowledge Graph as a Trainer} learn visual semantic embedding space supervised by a semantic embedding. They either learn a) a transformation function, e.g. MLP, on top of a pre-trained visual embedding space that suits as a transformation function or b) a visual-semantic features extractor that learns the final embedding directly.}
    \label{fig:Knowledge Graph as a Trainer overview}
\end{figure*}

Methods that belong to the category \emph{Knowledge Graph as a Trainer} combine the visual output of a DNN with the auxiliary knowledge of a KG by learning a visual-semantic embedding $h_{v,s}$.
Figure~\ref{fig:Knowledge Graph as a Trainer overview} illustrates a conceptual architecture of the knowledge graph as a trainer approach.
The KG acts as a trainer and supervises the training of the DNN using $h_s$, rather than letting the DNN learn a $h_v$ solely depending on the data distribution of the images.
We refer to such an embedding of visual information learned under the supervision of a semantic embedding $h_s$ as a visual-semantic embedding $h_{v,s}$.
To combine semantic and visual information, some approaches either learn a transformation function, e.g. MLP, on a pre-trained and fixed visual embedding $h_v$ or learn a visual-semantic features extractor $f_{v,s}(\cdot)$ directly.


\subsubsection{Visual-Semantic Transformation Models}
As shown in Figure~\ref{fig:Knowledge Graph as a Trainer MLP}, the pre-trained $h_{v}$ is fixed over the whole training process and an additional transformation function, e.g. MLP, is learned to transform $h_{v}$, into the visual-semantic embedding space $h_{v,s}$.

Akata et al.~\cite{DBLP:journals/pami/AkataPHS16} refer to their semantic embedding space transformations as label embedding methods.
They compared transformation functions from the visual embedding space to the attribute label embedding space, the hierarchy label embedding space, and the Word2Vec~\cite{DBLP:conf/nips/MikolovSCCD13} label embedding space.
Lonij et al.\cite{DBLP:journals/corr/abs-1708-08310} approached the task of open-world visual recognition by using KGs.
They learn $h_s$ from a WordNet KG by using the \emph{neural tensor layer} (NTL)~\cite{DBLP:conf/nips/SocherCMN13} architecture and embedd the visual embedding generated by a pre-trained CNN into the same space using the hinge rank loss.

\paragraph{Related Approaches using other Auxiliary Knowledge:}
One of the first approaches that use semantic embeddings with NNs is the work from Mitchell et al.~\cite{Mitchell1191}. 
They use language embeddings derived from text corpus statistics to generate neural activity pattern images.
Instead of generating images from text, Palatucci et al.~\cite{DBLP:conf/nips/PalatucciPHM09} learn a linear regression model to map neural activity patterns into language embedding space.
Socher et al.~\cite{DBLP:conf/nips/SocherGMN13} present a model for zero-shot learning that learns a transformation function between a visual embedding space, obtained by an unsupervised feature extraction method, and a semantic embedding space, based on a language model.
The authors trained a 2-layer NN with the MSE loss to transform the visual embedding into the language embedding of 8 classes.
Frome et al.~\cite{DBLP:conf/nips/FromeCSBDRM13} introduce the deep visual-semantic embedding model DeViSE that extends the approach from 8 known and 2 unknown classes to 1,000 known and 20,000 unknown classes.
Therefore, they pre-train their visual features extractor using ImageNet and their semantic embedding vector using a skip-gram language model~\cite{DBLP:conf/nips/MikolovSCCD13}.
In contrast to Socher et al.~\cite{DBLP:conf/nips/SocherGMN13} they learn a linear transformation function between the visual embedding space and the semantic embedding space using a combination of dot-product similarity and hinge rank loss since they claim that MSE distance fails in high dimensional space.
Norouzi et al.~\cite{DBLP:journals/corr/NorouziMBSSFCD13} propose \emph{convex combination of semantic embeddings} (ConSE).
ConSE performs a convex combination of known classes in the semantic embedding space, weighted by their predicted output scores of the DNN, to predict unknown classes in a zero-shot learning task.
Similarly, Zhang et al.~\cite{DBLP:conf/iccv/ZhangS15a} introduce the \emph{semantic similarity embedding} (SSE), which models target data instances as a mixture of seen class proportions.
They built a semantic space that each novel class could be represented as a probabilistic mixture of the projected source attribute vectors of the known classes.

Kodirov et al.~\cite{DBLP:conf/cvpr/KodirovXG17} propose SAE a semantic autoencoder for zero-shot learning.
It is learned by encoding pre-trained visual features of a CNN into a latent semantic space and then by decoding them back into visual space.
The semantic space is based on class attributes for smaller datasets and on a word2vec language model for larger datasets.
They claim that their latent semantic embedding space can better handle the projection domain shift problem, i.e. the distribution shift between seen and unseen classes.





\subsubsection{Visual-semantic Features Extractors}

As illustrated in Figure~\ref{fig:Knowledge Graph as a Trainer} the visual-semantic features extractor $f_{v,s}(\cdot)$ is learned to directly transform the images into a visual-semantic embedding $h_{v,s}$ using the supervision of the semantic embedding space $h_s$.
As described in Section~\ref{ssec:Knowledge Graph Embedding}, $h_s$ is mostly learned using an unsupervised KGE-Method and $f_{v,s}(\cdot)$ is implemented using a standard DNN.


Monka et. al~\cite{DBLP:conf/semweb/MonkaH0R21} propose KG-NN, an approach that uses a KG and its $h_s$ to train a visual DNN.
Using a contrastive knowledge graph embedding loss in combination with $h_s$ they learn a visual-semantic features extractor $f_{v,s}(\cdot)$.
They test their approach on domain generalization and adaptation tasks for road sign recognition in Germany and China, as well as on mini-ImageNet and various derivatives.
They show that their visual features extractor learned using the \emph{Knowledge Graph as a Trainer} outperforms a conventional DNN trained with CE, the same DNN without additional information from the KG, and the same DNN using additional information from a pre-trained GloVe embedding in visual transfer learning tasks.

Jayathilaka et al.~\cite{DBLP:conf/pkdd/JayathilakaMS21} proposed a framework named ViOCE that integrates ontology-based background knowledge in the form of n-ball class embeddings into a DNN-based vision architecture.
The approach consists of two components - converting symbolic knowledge of an ontology into continuous space by learning n-ball embeddings that capture properties of subsumption and disjointness and guiding the training and inference of a vision model using the learned embeddings.

\paragraph{Related Approaches using other Auxiliary Knowledge:}
Joulin et al.~\cite{DBLP:conf/eccv/JoulinMJV16} demonstrate that feature extractors trained to predict words in image captions learn useful image representations. They convert the title, description, and hashtag metadata of images into a bag-of-words multi-label classification task and showed that pre-training a feature extractor to predict these labels learned representations which performed similarly to ImageNet-based pre-training on transfer tasks.
Radford et al.~\cite{radford2learning} claim that state-of-the-art CV systems are restricted to predict a fixed set of predetermined object categories.
Therefore, they propose to use a simple and general pre-training of their CNN with natural language supervision, i.e. predicting which caption goes with which image on a dataset of 400 million image-text pairs collected from the internet using the objective of Zhang et al.~\cite{DBLP:journals/corr/abs-2010-00747}.




\subsection{Knowledge Graph as a Peer}
\label{ssec:Knowledge Graph as a Peer}
\begin{figure*}[t]
    \centering
        \begin{subfigure}[t]{0.49\textwidth}
    \centering
    \includegraphics[width=\textwidth]{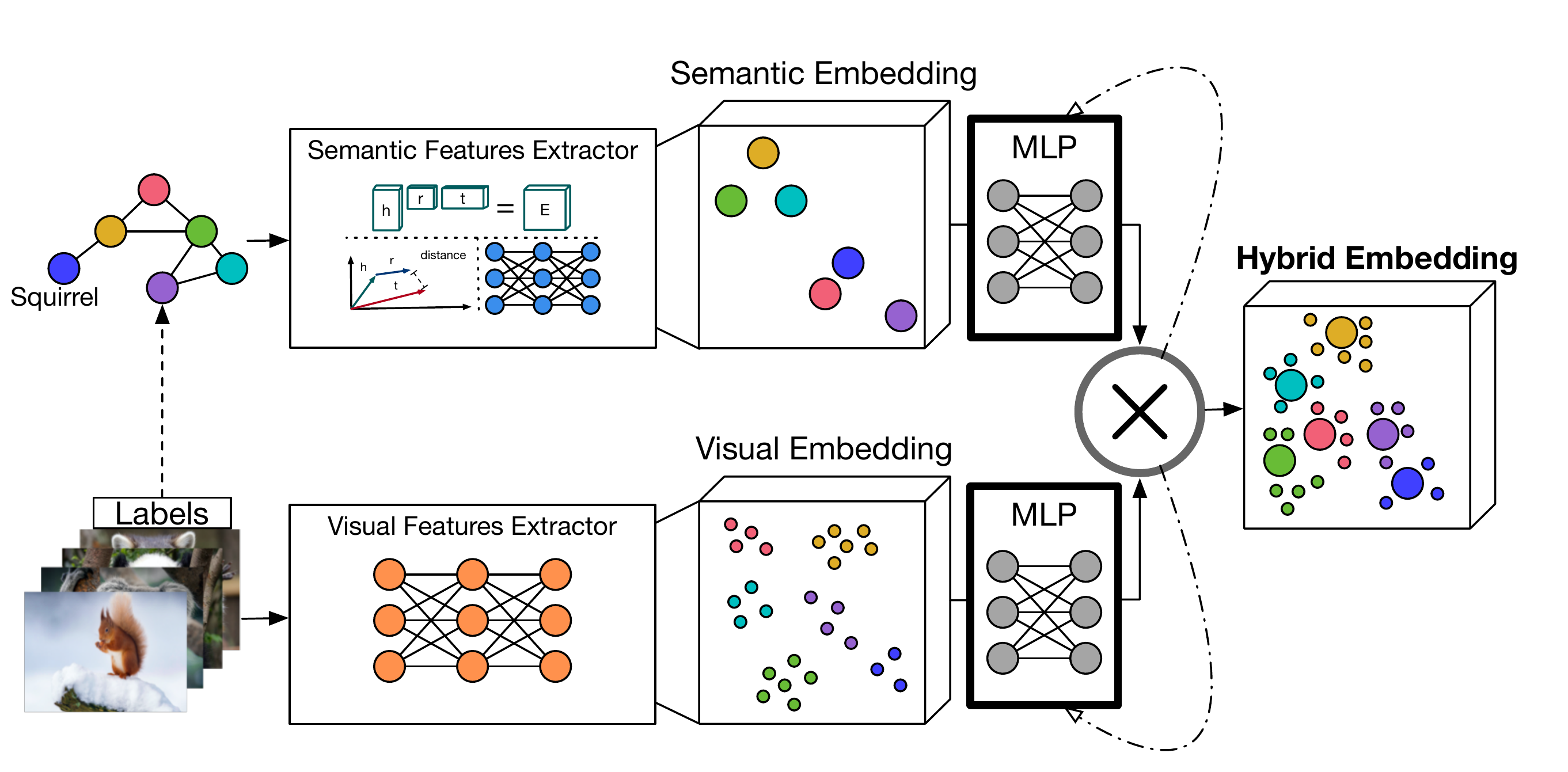}
    \caption{Hybrid transformation model}
    \label{fig:Knowledge Graph as a Peer MLP}
    \end{subfigure}
    ~
    \begin{subfigure}[t]{0.49\textwidth}
    \centering
    \includegraphics[width=\textwidth]{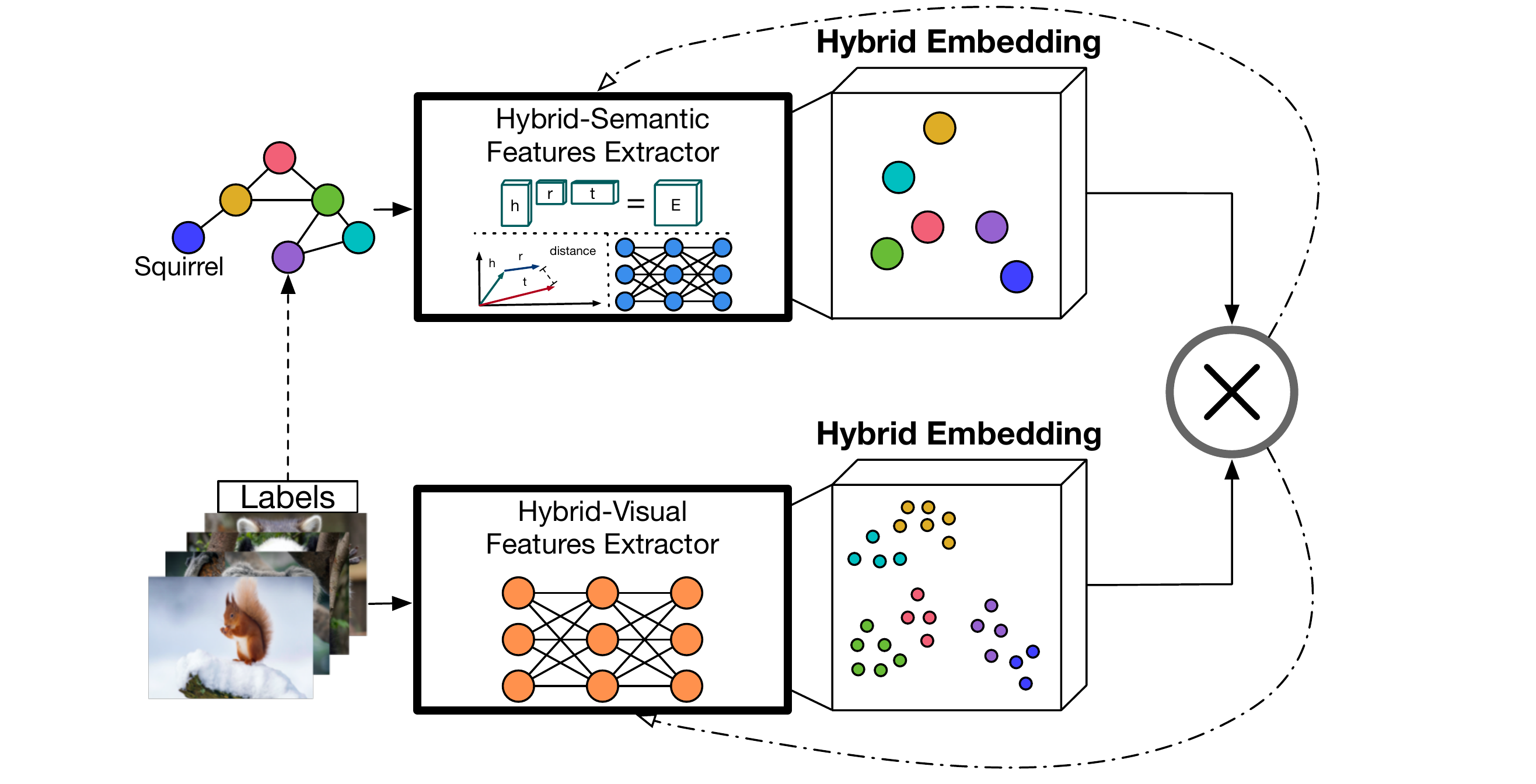}
    \caption{Hybrid features extractor}
    \label{fig:Knowledge Graph as a Peer}
    \end{subfigure}
    \caption{Approaches that belong to the category \emph{Knowledge Graph as a Peer} learn hybrid embedding space as a combination of visual and semantic embedding space. They either learn a) transformation functions, e.g. MLPs, on top of both pre-trained visual and semantic embedding spaces that suit as a transformation function or b) hybrid features extractors that learn the final embedding directly.}
    \label{fig:Knowledge Graph as a Peer overview}
\end{figure*}

Approaches of the category \emph{Knowledge Graph as a Peer} combine the visual DNN with the auxiliary knowledge of a KG by influencing both semantic and visual embedding.
Unlike the previous categories, the idea of a hybrid embedding $h_h$ is to fuse the visual embedding $h_v$ and the semantic embedding $h_s$.
Both semantic and visual data are then embedded into $h_{h}$.
Figure~\ref{fig:Knowledge Graph as a Peer overview} illustrates a conceptual architecture of the knowledge graph as a peer approach.
The final hybrid embedding space is either a combination of pre-trained visual embedding $h_v$ and semantic embedding $h_s$, using a transformation function, e.g. MLP, or a combination of hybrid-visual $f_{h,v}(\cdot)$ and hybrid-semantic features extractors $f_{h,s}(\cdot)$.

\subsubsection{Hybrid Transformation Models}
As shown in Figure~\ref{fig:Knowledge Graph as a Peer MLP}, pre-trained $h_{s}$ and pre-trained $h_v$ are fixed over the whole training process and an additional transformation functions, e.g. MLPs, are learned to transform $h_{s}$ and $h_v$, into the hybrid embedding space $h_{h}$.

Zhao et al.~\cite{DBLP:conf/iccv/ZhaoPZF017} propose a joint model that combines an image stream and a concept stream via a joint loss function to preserve concept hierarchy as well as visual feature similarities.
The concept stream is based on a language embedding with the hierarchical graph of WordNet and the image stream is a visual embedding from semantic segmentation DNN.
They compare their approach against the standard CE-based approach and semantic embedding space transformations based on Word2Vec.
Roy et al.~\cite{DBLP:journals/corr/abs-2012-06236} introduce a zero-shot learning model that takes advantage of the commonsense knowledge graph ConceptNet 5.5 to generate $h_s$ of the class labels by using a GCN-based autoencoder.
They enrich $h_s$ with additional attributes and language embeddings, which is then compared with a pre-trained visual output of a DNN using a relation network~\cite{DBLP:conf/cvpr/SungYZXTH18}.

\paragraph{Related Approaches using other Auxiliary Knowledge:}
Yang et al.~\cite{DBLP:journals/corr/YangH14a} propose a two-sided NN to learn a combination of a pre-trained visual embedding and a semantic embedding of attributes and word vectors based on image descriptions to perform zero-shot learning and domain generalization.
To train their NN they use a Euclidean loss for regression and a hinge rank loss for classification.
Fu et al.~\cite{DBLP:journals/pami/FuHXG15} try to reduce the bias of semantic embedding spaces, by proposing a transductive multi-view embedding framework that aligns novel features with the semantic embedding space for zero-shot learning.
The framework first transforms the semantic embedding space into a joint embedding space using the unlabeled target data with a multi-view \emph{canonical correlation analysis} (CCA) to alleviate the projection domain shift problem.
And Second, a heterogeneous multi-view hypergraph label propagation method is used to perform zero-shot learning in the transductive embedding space, which combines additional semantic knowledge in the form of attributes and word vectors from related classes.
Ba et al.~\cite{DBLP:conf/iccv/BaSFS15} introduce a flexible zero-shot learning model that learns to predict unseen image classes using a language embedding.
Therefore, they add two separate MLPs on top of the visual embedding and the semantic embedding and train them using the binary-CE loss, the hinge loss, and the Euclidean distance loss.
Karpathy et al.~\cite{DBLP:conf/cvpr/KarpathyL15} learn a model that generates language descriptions for detected objects in an image.
Their objective aligns the output of a pre-trained CNN applied to image regions, and the output of a bidirectional RNN applied to sentences.
Changpinyo et al.~\cite{DBLP:conf/cvpr/ChangpinyoCGS16} use a set of “phantom” object classes whose coordinates live in both the semantic space and the model space.
To align the two spaces, they view the coordinates in the visual embedding as the projection of the vertices on the graph from the semantic embedding.
To compute low-dimensional Euclidean space embeddings from the weighted graph they propose to use the algorithm of Laplacian eigenmaps, mapping semantic and visual embedding into a common space defined by the mixture of seen classes proportions.
Tsai et al.~\cite{DBLP:conf/iccv/TsaiHS17} propose the approach ReViSE that learns an unsupervised joint embedding of semantic and visual features to enable zero-shot learning.
As external knowledge, they experiment with three different embedding methods for their attributes, human-annotated attributes~\cite{DBLP:journals/pami/LampertNH14}, Word2Vec attributes, and GloVe attributes.
Tang et al.~\cite{DBLP:journals/pami/TangWWGDGC18} propose the \emph{large scale detection through adaptation} (LSDA) framework to improve object detectors with image classification DNNs, hence without requiring expensive bounding box annotations.
LSDA defines visual similarity as the distance between pre-trained visual embedding vectors and semantic similarity as the distance between pre-trained language embedding vectors of the labels.
Jiang et al.~\cite{DBLP:conf/iccv/Jiang0SC19} introduce their \emph{transferable contrastive network} (TCN) explicitly transfers knowledge from the source classes to the target classes, to counteract the overfitting problem on source classes.
To compute the similarities between classes in the hybrid embedding space, they design a contrastive network that automatically judges how well the embedding vector is consistent with a specific class.
Li et al.~\cite{DBLP:conf/eccv/Li0LZHZWH0WCG20} propose a multi-layer transformer~\cite{DBLP:conf/nips/VaswaniSPUJGKP17} model as DNN, which uses object tags detected in images as anchor points to learn a joint embedding of the detected objects and the language tags, instead of simply concatenating visual embedding and semantic embedding.
Yu et al.~\cite{DBLP:journals/corr/abs-2006-16934} propose a knowledge-enhanced approach, ERNIE-ViL, to learn joint representations of vision and language using a transformer model as DNN.
ERNIE-ViL tries to construct the detailed semantic connections across vision and language while constructing a scene graph parsed from sentences and type prediction tasks, i.e., object prediction, attribute prediction, and relationship prediction in the pre-training phase.




\subsubsection{Hybrid Features Extractors} 

As depicted in Figure~\ref{fig:Knowledge Graph as a Peer}, hybrid-semantic $f_{h,s}(\cdot)$ and hybrid-visual $f_{h,v}(\cdot)$ features extractors are learned to directly transform KG and images into a common hybrid embedding $h_{h}$.
As described in Section~\ref{ssec:Knowledge Graph Embedding}, $f_{h,s}(\cdot)$ is usually implemented using a supervised KGE-Method and $f_{h,v}(\cdot)$ using a standard DNN.

Recently, Naeem et. al~\cite{DBLP:conf/cvpr/NaeemXTA21} proposed a method to perform zero-shot image classification using hybrid features extractors.
An ImageNet pre-trained DNN is used for the visual features extractor and a GCN in the \emph{compositional graph embedding} (CGE) setting is used for the semantic features extractor.
However, they learn a joint embedding function that can influence the weights of the DNN as well as the weights from the GCN.
Interestingly, they compare their model against a similar version of their model, but with a fixed visual features extractor where the KG just acts as a trainee (see Section~\ref{ssec:Knowledge Graph as a Trainee}).
They use that version for comparison with related approaches, stating that all other methods are based on fixed visual features extractors.
Moreover, they show that a hybrid approach with an adaptive visual features extractor performs better than the other.

\paragraph{Related Approaches using other Auxiliary Knowledge:}
Zhang et al.~\cite{DBLP:journals/corr/abs-2010-00747} use two contrastive pre-training objectives, contrasting semantic embedding to visual embedding, and vice versa, on the special domain of medical imaging to learn a joint feature extractor. 
Instead of previous works that learn transformation functions on top of fixed image trained visual features extractors they directly supervise the training of the CNNs with language embedding information.
To train their DNN they use text-image paired data.



\section{Visual Transfer Learning Datasets and Benchmarks}
\label{sec:Datasets and Benchmarks}
Building expressive knowledge graphs from scratch can be a quite challenging task.
Concerning \textbf{RQ3}, this section provides an overview of standard and large-scale KGs that can be used as auxiliary knowledge.
Moreover, as there are no standard datasets and benchmarks to compare visual transfer learning tasks that use KGs, we refer to \textbf{RQ4} and provide a list of datasets and benchmarks that have been used in the community of knowledge-based ML and visual transfer learning in Table~\ref{tab:Datasets and Benchmarks}.
These Datasets and Benchmarks include: 
a) Attribute augmented image datasets with textual image or class attribute descriptions;
b) Language augmented image datasets, providing additional textual descriptions of the images;
c) Knowledge graph augmented image datasets, containing meta information of class relations in a KG;
d) Image datasets without auxiliary knowledge, used for zero-shot learning and domain generalization tasks.

\begin{table*}[htb]
    \centering
    \begin{tabular}{|c|c|c|c|c|c|c|}
    \hline
        \textbf{Type of Knowledge} & \textbf{Task} & \textbf{Dataset} & \textbf{Auxiliary Knowledge} & \textbf{Release Date}\\
        \hline
        \multirow{5}{*}{\begin{tabular}{c}
        Attributes  \\
        + \\
        Images \\      
        \end{tabular}} & \multirow{4}{*}{ZSL} & AwA & textual attributes for img/cls  & 2009\\
        \cline{3-5}
        & & AwA2 &  textual attributes for img/cls &  2019\\
        \cline{3-5}
        & & SUN &  textual attributes for img/cls &  2012\\
        \cline{3-5}
        & & CUB &   textual attributes for img/cls & 2010\\
        \cline{2-5}
        & DG & Large-Scale Car Dataset & textual attributes for img/cls  & 2017\\
        \hline
        \multirow{4}{*}{\begin{tabular}{c}
        Language  \\
        + \\
        Images \\      
        \end{tabular}}
        & \multirow{4}{*}{Other} & MS-COCO &   textual denotation graph &  2014\\
        \cline{3-5}
        & & Flickr30K &   textual denotation graph & 2015 \\
        \cline{3-5}
        & & SBU Captions &   textual descriptions for img & 2011 \\
        \cline{3-5}
        & & Conceptual Captions &  textual descriptions for img  & 2018 \\
        \hline
        \multirow{4}{*}{\begin{tabular}{c}
        Knowledge Graph  \\
        + \\
        Images \\      
        \end{tabular}}& \multirow{3}{*}{ZSL} 
        & Visual Genome &  flat concept graph &  2017 \\
        \cline{3-5}
        & & miniImageNet &  hierarchical concept graph 
        & 2016 \\
        \cline{3-5}
        & & tiredImageNet &  hierarchical concept graph
        & 2018 \\
        \cline{2-5}
        & DG & ImageNet &  hierarchical concept graph & 2009-2015\\
        \hline
        \multirow{5}{*}{Images}& 
        \multirow{2}{*}{ZSL} & CIFAR-FS &  N/A &   2016 \\
        \cline{3-5}
        & & FC-100 &  N/A &   2016 \\
        \cline{2-5}
        & \multirow{3}{*}{DG} & Office-31 &   N/A & 2010\\
        \cline{3-5}
        & & Office-Home &  N/A &   2016 \\
        \cline{3-5}
        & & VisDA2017 &  N/A &   2017 \\
        \hline
    \end{tabular}
    \caption{Datasets and benchmarks of the field of visual transfer learning and knowledge-based ML are summarized due to type of knowledge, task, auxiliary knowledge, and their release date. ZSL is zero-shot-learning, DG is domain generalization, and other are tasks from image classification, object detection, object segmentation, and image captioning.}
    \label{tab:Datasets and Benchmarks}
\end{table*}

\subsection{Generic Knowledge Graphs}
\label{ssec:Knowledge Graphs}
Over the years, several open-access KGs have been created by various community initiatives. 
These graphs contain universal knowledge which potentially can be used as auxiliary knowledge in various scenarios.
In the following, some of the most common generic KGs currently available are described in more detail.
However, for deeper insights, we refer to the survey of Färber et al.~\cite{farber2015comparative}.

\paragraph{WordNet~\cite{DBLP:journals/cacm/Miller95}:}
WordNet, firstly released in 1995, is an online lexical reference system for English nouns, verbs, and adjectives which are organized into \emph{synonym sets} (synsets), each representing one underlying lexical concept.
WordNet superficially resembles a thesaurus, in that it groups words based on their meanings.
There are 117,000 synsets, each synset is linked with other synsets by super-subordinate relations, forming a hierarchical structure of instances, concepts and categories whereas all are linked with the root node, \emph{entity}.

\paragraph{ConceptNet 5.5~\cite{DBLP:conf/aaai/SpeerCH17}:}
ConceptNet 5.5 is a KG that connects words and phrases of natural language with labeled edges.
Its knowledge is collected from many sources that include expert-created resources, crowd-sourcing, and games with a purpose. 
It is designed to represent the general knowledge involved in understanding language, improving natural language applications by allowing the application to better understand the meanings behind the words people use.
Information within ConceptNet is modeled as a directed labeled graph (see Section~\ref{ssec:Knowledge Graph}), where concepts are connected via binary relationships.
It contains approximately 34 million statements, i.e. edges~\footnote{https://conceptnet.io}.

\paragraph{DBPedia~\cite{DBLP:conf/semweb/AuerBKLCI07}:}
DBPedia is a community effort to extract structured information from Wikipedia and to make this information available on the Web.
DBpedia allows you to ask sophisticated queries against datasets derived from Wikipedia and to link other datasets on the Web to Wikipedia data.
The underlying structure of DBpedia is a hypergraph model (see Section~\ref{ssec:Knowledge Graph}) where facts are represented via binary and n-ary relationships.  
The English version of the DBpedia knowledge base describes 4,58 million things, out of which 4,22 million are classified in a consistent ontology, including 1,445,000 persons, 735,000 places, and 411,000 creative works~\footnote{https://wiki.dbpedia.org/about}.

\paragraph{Wikidata~\cite{DBLP:conf/www/Vrandecic12}:} 
Wikidata is a KG, built collaboratively by humans or automated agents.
It encapsulates facts about the world entities organized in a form of complex statements. 
The basic structure comprises items defined with a label and several aliases. 
In addition, Wikidata contains some sense of basic commonsense knowledge~\cite{DBLP:journals/corr/abs-2008-08114} which allows for performing several sophisticated downstream tasks based on reasoning capabilities. 
The facts within Wikidata are represented as a hyper-relation graph (see Section~\ref{ssec:Knowledge Graph}) where relations are enriched with additional information known as qualifiers~\cite{DBLP:conf/emnlp/GalkinTMUL20}.
These qualifiers enable the disambiguation of complex facts about the same entities in different contexts.
Currently, Wikidata has 92,4 million items, where around 6,3 million of them are humans, 2 million administrative entities, 22,5 million scholarly articles, and so on~\footnote{\url{https://www.wikidata.org/wiki/Wikidata:Statistics}, accessed on 02 February 2021}.

\subsection{Image Datasets with Auxiliary Knowledge}
\label{Datasets with Auxiliary Knowledge}
Some datasets are built on auxiliary knowledge bases or intended to use with auxiliary information.
We provide a categorization of the datasets and benchmarks concerning the type of auxiliary knowledge it is augmented with.

\subsubsection{Attribute Augmented Image Datasets}
\label{sssec:Attribute Augmented Image Datasets}
Attribute augmented image datasets are image datasets with additional descriptions of image and class attributes, used for knowledge-based ML.

\paragraph{AwA~\cite{DBLP:conf/cvpr/LampertNH09}:}
The \emph{Animals with Attributes} dataset consists of over 30,000 images with pre-computed reference features for 50 animal classes, for which a semantic attribute annotation is available from studies in cognitive science.
However, as the AWA images do not have a public copyright license, only some computed image features, i.e. SIFT~\cite{DBLP:journals/ijcv/Lowe04}, DECAF~\cite{DBLP:conf/icml/DonahueJVHZTD14}, VGG19~\cite{DBLP:journals/corr/SimonyanZ14a} of
AWA dataset are publicly available, rather than the raw images.
Since image feature learning is an important part of modern CV, this dataset is of limited use for end-to-end learned visual models.

\paragraph{AwA2~\cite{DBLP:journals/pami/XianLSA19}:}
The \emph{Animals with Attributes 2} dataset is recently introduced and has roughly the same number of images all with public licenses, and the same number of classes and attributes as the AwA dataset.

\paragraph{CUB~\cite{WelinderEtal2010}:}
The \emph{Caltech-UCSD-Birds 200-2011} dataset is a fine-grained and medium scale dataset concerning both the number of images and the number of classes, i.e. 11,788 images from 200 different types of birds annotated with 312 attributes. Akata et al.~\cite{DBLP:journals/pami/AkataPHS16} introduces the first zero-shot split of CUB with 150 training, 50 validation, and 50 test classes.

\paragraph{SUN~\cite{DBLP:conf/cvpr/PattersonH12}:}
The \emph{Scene Categorization Benchmark} is also a fine-grained and medium-sized dataset, both in terms of the number of images and the number of classes., i.e. SUN contains 14,340 images coming from 717 types of scenes annotated with 102 attributes.
Lampert et al.~\cite{DBLP:journals/pami/LampertNH14} use 645 classes of SUN for training, 65 classes for validation, and 72 classes for testing.

\paragraph{Large-Scale Car Dataset~\cite{DBLP:conf/aaai/GebruKWCDF17}:}
The \emph{Large-Scale Car Dataset} originally consists of 2,657 classes and 1,095,021 images from four sources: craigslist.com, cars.com, edmunds.com and Google Street View. They refer to images from craigslist.com, cars.com and edmunds.com as web images and those from Google Street View as GSV images.
It was adapted to domain generalization using a subset of 170 classes and 71,030 images~\cite{DBLP:conf/iccv/GebruHF17}.
The image category web images is used as source domain, whereas the category GSV images suits as target domain.
The cars in web images are large and typically un-occluded whereas those in GSV are small, blurry and occluded.
In addition to the category labels, each class is accompanied by metadata such as the make, model body type, and manufacturing country of the car.

\subsubsection{Language Augmented Image Datasets}
\label{sssec:Language Augmented Image Datasets}
These image datasets are enriched with additional textual descriptions and captions of images.
To categorize images based on the textual descriptions, denotation graphs are introduced and are available for some datasets.
\paragraph{MS-COCO~\cite{DBLP:conf/eccv/LinMBHPRDZ14}:}
\emph{MS-COCO} includes images of complex everyday scenes with common objects in their natural context. 
It contains a total of 2.5 million labeled instances of 91 object types, in 328k images, each accompanied with five human-written captions.
It is used for category detection, instance spotting, and instance segmentation.
Recently, Zhang et. al~\cite{DBLP:conf/emnlp/ZhangHJIS20} released an additionally learned denotation graph for MS-COCO, which induces a partial ordering over the textual image descriptions.
There is also work that extends \emph{MS-COCO} to zero-shot learning tasks by providing additional splits of unseen and seen class categories~\cite{DBLP:conf/eccv/BansalSSCD18}.

\paragraph{Flickr30K~\cite{DBLP:journals/tacl/YoungLHH14}:}
The \emph{Flickr30K} is a standard benchmark for sentence-based image description and was originally developed for the tasks of image-based and text-based retrieval.
The dataset contains 31K images collected from the Flickr website, with five textual descriptions per image.
Each image is described independently by five annotators who are not familiar with the specific entities and circumstances, resulting in high-level descriptions such as “Three people setting up a tent”. 
The images are under the Creative Commons license.
Moreover, they released a denotation graph for the dataset~\cite{DBLP:conf/emnlp/ZhangHJIS20}.

\paragraph{SBU Captions~\cite{DBLP:conf/nips/OrdonezKB11}:}
\emph{SBU Captions} contains a large number of images from the Flickr website.
They are filtered to produce a data collection containing over 1 million well-captioned images.
The images have rich user-associated captions from a web-scale captioned image collection.
These text descriptions generally work similarly to captions and usually relate directly to some aspect of the visual image content.

\paragraph{Conceptual Captions~\cite{DBLP:conf/acl/SoricutDSG18}:}
\emph{Conceptual Captions} consists of an order of magnitude more images than the MS-COCO dataset and represents a wider variety of both images and image caption styles.
Therefore, they extracted and filtered image caption annotations from billions of internet sources, e.g. webpages.

\subsubsection{Knowledge Graph Augmented Image Datasets}
\label{sssec:Knowledge Graph Augmented Image Datasets}
These datasets are augmented with an additional KG describing relations between classes or a scene in an image.
\paragraph{Visual Genome~\cite{DBLP:journals/ijcv/KrishnaZGJHKCKL17}:}
\emph{Visual Genome} provides a flat concept graph model of object relationships in images. 
Dense annotations of objects, attributes, and relationships within each image are collected. Specifically, the dataset contains over 100K images where each image has an average of 21 objects, 18 attributes, and 18 pairwise relationships between objects.
For zero-shot learning a split with 608 categories are considered for classification~\cite{DBLP:conf/eccv/BansalSSCD18,DBLP:conf/aaai/LuoZHY20}.
Among these, 478 are seen categories, and 130 are unseen categories. 
This results in 54,913 training images and 7,788 test images. 
The relationship graph in the dataset has 6,396 edges.

\paragraph{ImageNet~\cite{DBLP:journals/ijcv/RussakovskyDSKS15}:}
The \emph{ImageNet Large-Scale Visual Recognition Dataset and Challenge} is a benchmark in object category classification and detection on hundreds of categories and millions of images. 
The challenge has been run annually from 2010 to 2015.
It contains 1000 classes and more than 1,2 mil train, and 100K test images per class for object classification.
For object detection, it contains 1000 classes and more than 450K training images with 470K bounding boxes, 50K validation images with 55K bounding boxes, and 40K test images per class.

There are several derivatives of ImageNet with different appearances, as
\emph{ImageNetV2}~\cite{DBLP:conf/icml/RechtRSS19},
\emph{ImageNet Sketch}~\cite{DBLP:conf/aaai/WangLZYBXHW020},
\emph{ImageNet-Vid}~\cite{Shankar_2021_ICCV},
\emph{ImageNet Adversarial}~\cite{DBLP:conf/cvpr/HendrycksZBSS21},
\emph{ImageNet Rendition}~\cite{DBLP:journals/corr/abs-2006-16241},
and such with synthetic distribution shifts, as
\emph{ImageNet-C}~\cite{DBLP:conf/iclr/HendrycksD19},
and
\emph{Stylized ImageNet}~\cite{DBLP:journals/corr/abs-2004-07780}.
More recently, a domain generalization scenario has been created in which ImageNet-trained models are tested on various ImageNet derivatives to evaluate the robustness of the models to distribution shift.

\paragraph{MiniImageNet~\cite{DBLP:conf/nips/VinyalsBLKW16}:}
\emph{MiniImageNet} is a derivative of the ImageNet dataset and consisting of 60K color images of size 84 × 84 with 100 classes, each having 600 examples.
Since this dataset fits in memory on modern computers, it is very convenient for rapid prototyping and experimentation.
These 100 classes are divided into 64 train, 16 val, and 20 test classes for the zero-shot learning task.

\paragraph{TiredImageNet~\cite{DBLP:conf/iclr/RenTRSSTLZ18}:}
\emph{TiredImageNet} is a subset of the ImageNet dataset.
It groups classes into broader categories corresponding to higher-level nodes in the ImageNet hierarchy. There are 34 categories in total, with each category containing between 10 and 30 classes. 
For zero-shot learning they split the categories into 20 training, 6 validation, and 8 testing categories.
This ensures that all of the training classes are sufficiently distinct from the testing classes, unlike miniImageNet.

\subsection{Image Datasets without Auxiliary Knowledge}
This section introduces transfer learning image datasets that have been originally created without auxiliary knowledge.

\subsubsection{Zero-Shot Learning Datasets without Auxiliary Knowledge}
\label{sssec:Zero-Shot Datasets Without Auxiliary Knowledge}
We introduce image datasets that have been applied mainly for zero-shot learning or few-shot learning tasks.



\paragraph{CIFAR-FS~\cite{DBLP:conf/iclr/BertinettoHTV19}:}

\emph{CIFAR-FS} is randomly sampled from CIFAR-100~\cite{krizhevsky2009learning}.
CIFAR-100 contains 600 images in each of 100 classes, which are further grouped into 20 superclasses.
The limited original resolution of 32×32 makes the task harder and at the same time allows fast prototyping.
Moreover, the dataset is used for the task of few-shot learning.

\paragraph{FC100~\cite{DBLP:conf/nips/OreshkinLL18}:}
\emph{Fewshot-CIFAR100} is a derivative of the CIFAR-100 dataset and provides a few-shot learning split of the full CIFAR-100 dataset.
The dataset is split into superclasses, rather than into individual classes to minimize the information overlap. 
Thus the train split contains 60 classes belonging to 12 superclasses, the validation and test contain 20 classes belonging to 5 superclasses each.

\subsubsection{Domain Generalization Datasets without Auxiliary Knowledge}
\label{sssec:Domain Gerneralization Datasets Without Auxiliary Knowledge}
We provide a summary of image datasets that have been applied mainly for domain generalization or domain adaptation tasks.

\paragraph{Office-31~\cite{DBLP:conf/eccv/SaenkoKFD10}:}
\emph{Office-31} is an object recognition dataset which contains 31 categories and three domains, that is, \emph{Amazon} (A), \emph{Webcam} (W), and \emph{DSLR} (D). 
These three domains have 2817, 498, and 795 instances, respectively. The images in Amazon are the online e-commerce images taken from Amazon.com. 
The images in Webcam are the low-resolution images taken by web cameras. And the images in DSLR are the high-resolution images taken by DSLR cameras. 
In the experiments, every two of the three domains are selected as the source and the target domains, which results in six tasks.
The evaluation contains all 6 cross-domain tasks:
A→D, A→W, D→A, D→W , W→A,W→D.

\paragraph{Office-Home~\cite{DBLP:conf/cvpr/VenkateswaraECP17}:}
\emph{Office Home} contains 15,585 images of 65 categories, collected from 4 domains:
a) Art: 2421 artistic depictions of objects in the form of sketches, paintings, ornamentation, etc.; 
b) Clipart: a collection of 4379 clipart images; 
c) Product: 4428 images of objects without a background, akin to the Amazon category in Office dataset;
d) Real-World: 4357 images of objects captured with a regular camera.
The evaluation contains all 12 cross-domain tasks.

\paragraph{VisDA2017~\cite{DBLP:journals/corr/abs-1710-06924}:}
\emph{The 2017 Visual Domain Adaptation Dataset and Challenge} is focused on the simulation-to-reality shift and has two associated tasks: image classification and image segmentation. 
The goal in both tracks is to first train a model on simulated, synthetic data in the source domain and then adapt it to perform well on real image data in the unlabeled test domain.
VisDA2017 is the largest dataset for cross-domain object classification, with over 280K images across 12 categories in the combined training, validation, and testing domains. The image segmentation dataset is also large-scale with over 30K images across 18 categories in the three domains.

\section{Related Surveys}
\label{sec:Related Surveys}
Since our survey explores approaches that are at the intersection of visual transfer learning and knowledge-based machine learning, we look at well-known surveys from both fields in this section. Furthermore, we provide additional insight into surveys on the topic of explainable AI, as the field is strongly related to knowledge-based ML.

\paragraph{Visual Transfer Learning:}
Pan et al.~\cite{DBLP:journals/tkde/PanY10} and Zhang et al.~\cite{DBLP:journals/corr/abs-1903-04687} categorized the task of visual transfer learning into three main settings: inductive, transductive, and unsupervised transfer learning.
In inductive transfer learning the task changes from source to target, whereas the domain stays the same.
In transductive transfer learning, the source and target tasks are the same, while the source and target domains are different.
Finally, in the unsupervised transfer learning setting, similar to inductive transfer learning, the target task is different from but related to the source task.
However, unsupervised transfer learning focuses on solving learning tasks when no labeled data is available in the source and the target domain.
Weiss et al.\cite{DBLP:journals/jbd/WeissK016} separated the field into homogeneous and heterogeneous transfer learning, whereas approaches of the former are developed and proposed for handling the situations where the domains are of the same feature space and the latter refers to the knowledge transfer process in the situations where the domains have different feature spaces.
Kaboli et al.~\cite{kaboli:hal-01575126} reviewed and structured 20 transfer-learning approaches.
Wang et al.~\cite{DBLP:journals/ijon/WangD18} investigated the field from the domain change perspective. 
If the domain change is small they call it homogeneous transfer learning and if the domain change is large they call it heterogeneous transfer learning.
Zhang et al.~\cite{DBLP:journals/csur/ZhangLOX19} further separated the field of transfer learning into 17 different tasks, based on supervision, the amount of labeled data, and the size of the domain gap.
Zhang et al.~\cite{DBLP:journals/corr/abs-1903-04687} categorized transfer learning based on their adaptation process into weakly supervised learning, instance re-weighting, feature adaptation, classifier adaptation, deep network adaptation, and adversarial adaptation.
Wang et al.~\cite{DBLP:journals/tist/WangZYM19} provide a comprehensive survey about zero-shot learning methods and their different semantic spaces.
These semantic spaces can either be engineered semantic spaces, generated by attributes, lexicals, and text-keywords, or learned semantic spaces, as label-embeddings, text-embeddings, and image-representations.
Xian et al.~\cite{DBLP:journals/pami/XianLSA19} recently released a survey about zero-shot learning where they structured the field into methods that learn linear compatibility, nonlinear compatibility, intermediate attribute classifier, or hybrid models.

\paragraph{Knowledge-Based Machine Learning:}
Only a few surveys have investigated the field of knowledge-based ML.
Von Rueden et al.~\cite{DBLP:journals/corr/abs-1903-12394} recently published a survey about knowledge-based ML under the term \emph{informed machine learning}.
They structure the field based on the source of the knowledge, the representation of the knowledge, and the integration of the knowledge into the ML pipeline.
Further, Gouidis et al.~\cite{DBLP:conf/aaaiss/GouidisVPABP20} structured the knowledge-based ML literature into approaches with symbolic knowledge, commonsense knowledge, and the ability to learn new knowledge.
They give an overview of different works that combines ML with knowledge-based approaches in the field of CV.
They categorized the approaches due to their CV task, e.g. object detection, scene understanding, image classification,
their applied ML architecture, e.g. CNN, GNN, RCNN, and their loss function, e.g. scoring functions, probabilistic programming models, Bayesian Networks.
Ding et al.~\cite{DBLP:conf/iccms/DingYLW19} reviewed all ontology applications in the field of object recognition. 
Another research field in demand is \emph{Explainable AI}, where knowledge-based methods and ML approaches are combined.
Explainable AI refers to methods and techniques of ML such that the results of the solution can be understood by humans.
Futia et al.~\cite{DBLP:journals/information/FutiaV20} investigated the field of explainable AI using KGs and categorized approaches into knowledge matching, cross-disciplinary and interactive explanations.
Chen et al.~\cite{chen:hal-01934907} and Chari et al.~\cite{DBLP:series/ssw/ChariGSM20a} proposed to use hybrid explanations of a taxonomy generated for the end-user, including causal methods, neuro-symbolic AI systems, and representation techniques.
Seeliger et al.~\cite{DBLP:conf/semweb/SeeligerPK19} summarized semantic web technologies that can provide valid explanations for ML models, separating them due to their ML technique and semantic expressiveness.
Chen et al.~\cite{DBLP:conf/ijcai/ChenG0HPC21} recently proposed a survey about knowledge-aware zero-shot learning.
They divided the machine learning methods that approach the zero-shot learning task into three distinct categories: mapping function based, generative model based, and graph neural network based.
They provided an overview of different types of auxiliary knowledge, e.g. text, attribute, knowledge graph, and rule and ontology.

Aditya et al.~\cite{DBLP:conf/ijcai/AdityaYB19} provide a survey about reasoning mechanisms and knowledge integration methods for image understanding applications.

Besides an overview of frameworks that handle logic operations, they briefly discuss at which position auxiliary knowledge can be introduced into a DL pipeline:
i) Ahead of the DNN, through a pre-processing of domain knowledge and augmentation of training samples;
ii) Inside the DNN, through a vectorization of parts of the knowledge base and as an input to intermediate layers;
iii) Inside of the DNN, to inspire the neural network architecture; and
iv) After the DNN, as a post-processing using external knowledge.
We understand their taxonomy as a general explanation of where external knowledge can be induced into the DL pipeline.
For instance, our category \emph{Knowledge Graph as a Reviewer} is related to iv), since the KG can operate as a post-processing network on the output of the visual DNN.
However, we also see that the reasoning process of the \emph{Knowledge Graph as a Reviewer} can be applied on an intermediate visual feature layer of the DNN.
Similarly, the categories \emph{Knowledge Graph as a Trainee}, \emph{Knowledge Graph as a Trainer}, and \emph{Knowledge Graph as a Peer} have overlaps with categories ii) and iii).
However, in contrast to Aditya et al. our categories are described by the explicit information exchange between the visual and semantic embedding space.
Instead of a categorization based on the position of the knowledge induction, our categories depend on whether the semantic embedding inspires the visual embedding or vice versa.
Using our categories, we therefore describe four distinct principles used to combine the two modalities.

Our survey explores the field of visual transfer learning using KGs.
Rather than just structuring the field, we also aim to provide the necessary tools for using KGs with DL pipelines to facilitate a straightforward entry.
Therefore, we present different modeling structures for KGs, concepts about visual and semantic feature extractors, and different methods for converting KGs into a vector-based $h_s$.
The main contribution is a categorization into four distinct categories of how a KG can be used with a DL pipeline for visual transfer learning tasks.
To enable a fair comparison for approaches of visual transfer learning using KGs, we summarize available KGs, datasets, and benchmarks.

\section{Challenges and Open Issues}
\label{sec:Challenges and Open Issues}
Integrating auxiliary knowledge in form of a KG into the DL pipeline not only helps in tackling challenges such as catastrophic forgetting or the need for a huge amount of data in transfer learning scenarios, but it also improves the robustness of DL approaches against naturally occurring domain shift.
However, exploiting this type of knowledge brings up new challenges related to knowledge representation and utilization, which we are going to discuss in the following.

\paragraph{Relevant Knowledge and its Representation:}
A major challenging task when dealing with modeling the knowledge for a given domain is to analyze what type of knowledge is relevant for performing a given task.
Currently, the majority of approaches focus on exploiting only the type of knowledge that is truly irrelevant to the context.
Furthermore, the temporal aspects between pieces of knowledge are minimally exploited or not exploited at all.
As described in Section~\ref{ssec:Knowledge Graph}, various modeling structures exist that can be used to represent multidimensional information.
However, the difficulty raised here is keeping the trade-off between the relevant knowledge and complexity of structures used to represent that. 

\paragraph{Evolving Knowledge:}
In daily scenarios, CV-related applications based on ML consume an abundant amount of data collected from various sensors.
Typically, this information is used for training purposes in form of vectors performing complex calculations to learn mathematical functions that best fit downstream tasks.
A crucial challenge here is to extract and integrate heterogeneous knowledge that can be managed and refined by humans.
Progress in the field of KG construction by embedding methods of language and information extraction has already been achieved.~\cite{DBLP:conf/www/DimouSCVMW14,DBLP:journals/ieicet/Kertkeidkachorn18,DBLP:journals/fgcs/DessiORBM21}.
This would enable the definition of different complex rules and reusable knowledge structures which later can be incorporated back to the existing or new ML pipelines.

\paragraph{Knowledge Embedding Methods:}
As we pointed out in Section~\ref{sssec:semantic features extractor}, there is a strong relation between knowledge graph embeddings and language embeddings as both are generated by a semantic feature extractor.
Using this assumption, we can apply knowledge graph embeddings in various new domains, where language embeddings have shown great potential, with the advantage that $h_s$ can be manually adapted to our needs.
This is done either by refining the knowledge in a KG or by using a particular embedding method relevant to the graph structure to best represent the inherent knowledge.
The challenge here is related to find suitable KGs and their modeling techniques to form either task-specific or universal $h_s$ spaces that support and enhance DL approaches in CV.

\paragraph{Joint Embedding Learning:}
We have seen that basic supervised learning methods that use CE tend to overfit the training data, leading to extensive problems when applied scenarios with a domain shift.
Finding a good embedding space is crucial which would enable it to be applied to multiple downstream tasks.
To learn efficiently on high dimensional spaces, energy-based functions instead of maximum likelihood seem to be promising, which should be further investigated under different requirements, like imbalance distribution within datasets. 
As described in Section~\ref{sssec:Training objectives}, the quality of the combination of visual and semantic embedding space is highly dependent on the similarity measure, the training objective, and the optimization method.
It is still an open challenge how to best fit these three parameters to find accurate combinations for a joint embedding space.
Moreover, learning visual features extractors directly on semantic embedding spaces with other features, e.g., temporal or contextualized embeddings, instead of discrete labels is a major challenge for future research.

\section{Discussion and Conclusion}
\label{sec:conclusion}
Visual transfer learning using different types of auxiliary knowledge has gained increasing attention in research.
Since initiatives for building and maintaining generic knowledge graphs host a large research community, we believe that exploiting them with DL will improve various applications, especially in visual transfer learning.
The insights gained in this survey can be useful to conceive solutions for addressing the identified challenges and open issues.

The survey investigates various forms of how KGs as a unified representation of auxiliary knowledge can be used based on a deep analysis of existing approaches.
Different graph models, corresponding embedding methods, and suitable training objectives to operate on high-dimensional spaces are described in detail.
The major contributions of the survey are formulated in four research questions presented in Section~\ref{sec:Methodology}.
The answers to these questions are given as follows:
\begin{itemize}
    \item \textbf{RQ1} - \emph{How can a knowledge graph be combined with a deep learning pipeline?}
    
    Approaches of the field of visual transfer learning using KG can be separated into four distinct categories based on how the KG is combined with the DL pipeline:
    
    1) \emph{Knowledge Graph as a Reviewer} - where the KG is used for post-validation of a visual model; 
    
    2) \emph{Knowledge Graph as a Trainee}, where a semantic-visual embedding $h_{s,v}$ is learned using a visual embedding $h_v$ as objective;
    
    3) \emph{Knowledge Graph as a Trainer}, a visual-semantic embedding $h_{v,s}$ is learned using a semantic embedding $h_s$ as objective; and 
    
    4) \emph{Knowledge Graph as a Peer}, where a hybrid-embedding $h_h$ is learned using a combination of semantic embedding $h_s$ and a visual embedding $h_v$ as objective.
    
    
    
    
    
    \item \textbf{RQ2} - \emph{What are the properties of the respective combinations?}
    It can be seen that every category has its applications in distinct tasks.
    
    1) \emph{Knowledge Graph as a Reviewer} - approaches leverage auxiliary knowledge by using it as an independent post-validation.
    The KG or $h_s$ enables reasoning over the output or intermediate feature layers of the DNN.
    However, the modalities are either learned independently or in sequential order, so that semantic and visual embedding space are not directly influenced by each other.
    
    2) \emph{Knowledge Graph as a Trainee} - approaches leverage auxiliary knowledge by providing a structure for a KGE-Method, e.g. GNN, that is learned using $h_v$ as objective.
    Approaches are used mainly in the zero-shot learning scenario to extend the learned model to classes that are not present in the training data, using the inductive property of GNNs combined with the ability of DNNs to extract relevant features of images. 
    
    3) \emph{Knowledge Graph as a Trainer} - approaches leverage auxiliary knowledge by influencing DNNs in learning specific visual features.
    The DNN can learn an image data distribution independent embedding provided by $h_s$ instead of just using the data distribution.
    Thus, we see the advantage of these approaches specifically in the domain generalization scenario. 
    
    4) \emph{Knowledge Graph as a Peer} - approaches leverage auxiliary knowledge by influencing semantic and visual embedding equally.
    Although it is not clear which modality dominates the other and therefore the learned embedding, approaches have yielded quite promising results for zero-shot learning and domain generalization tasks.
    
    \item \textbf{RQ3} - \emph{Which knowledge graphs already exist, that can be used as auxiliary knowledge?}
    We provide a short overview of generic KGs that could be used as a basis to form either specific or general approaches for the task of visual transfer learning using KGs.
    
    \emph{WordNet}, an online lexical reference system for English nouns, verbs, and adjectives, often used to build hierarchical relationship graphs of classes in the image dataset.
    
    \emph{ConceptNet 5.5}, a commonsense KG that connects words and phrases of natural language, often used to provide flat relationships between different classes of the image dataset.
    
    \emph{DBPedia}, a KG that represents structured information from Wikipedia and therefore allows to extract facts.
    
    \emph{Wikidata}, a commonsense KG built collaboratively by humans or automated agents with reasoning capabilities.
    
    \item \textbf{RQ4} - \emph{What datasets exist, that can be used in the combination with auxiliary knowledge to evaluate visual transfer learning?}
    We present several vision datasets and cluster them based on the type of auxiliary data they are augmented with.
    
    \emph{Attribute Augmented Image Datasets}, as Awa, Awa2, CUB, SUN, and Large-Scale Car Dataset.
    
    \emph{Language Augmented Image Datasets}, as MS-COCO, Flickr30K, SBU Captions, and Conceptual Captions.
    
    \emph{Knowledge Graph Augmented Image Datasets}, as Visual Genome, ImageNet, miniImageNet, and tiredImageNet.

    \emph{Image Datasets without Auxiliary Knowledge} for zero-shot learning, as CIFAR-FS, FC100, or domain generalization, as Office-31, Office-Home, and VisDA2017.
    
\end{itemize}

Future work is directed on conducting extensive experiments using KGs for visual transfer learning tasks while measuring various metrics, such as precision, recall, and accuracy.
Furthermore, it will be relevant to investigate the impact of knowledge structures represented via the three common graph models, the impact of different KGE-Methods, and the impact of the four categories a KG can be combined with the DL pipeline on the metrics as above.
We hope that this survey will help the reader to combine the technology of KGs and DL to develop models that can benefit from the appropriate combination of visual information with underlying semantic information.

\section{Acknowledgement}

This publication was created as part of the research project "KI Delta Learning" (project number: 19A19013D) funded by the Federal Ministry for Economic Affairs and Energy (BMWi) on the basis of a decision by the German Bundestag.


\bibliographystyle{ios1}           
\bibliography{bibliography}        

%

\end{document}